\documentclass{article}

\usepackage[numbers]{natbib}

\usepackage[preprint]{neurips_2020}

\usepackage{amsmath,amsfonts,amssymb}
\usepackage{url}
\usepackage{graphicx}
\usepackage{subcaption}
\usepackage{bbm}
\usepackage{algorithmic}
\newcommand{\bx}{\textbf{x}}
\newcommand{\bX}{\textbf{X}}

\newcommand{\by}{\textbf{y}}
\newcommand{\bK}{\textbf{K}}
\newcommand{\bA}{\textbf{A}}
\newcommand{\bB}{\textbf{B}}
\newcommand{\bS}{\mathbf{\Sigma}}
\newcommand{\bL}{\textbf{L}}
\newcommand{\bKmu}{\textbf{K}^{f}}
\newcommand{\bKsigma}{\textbf{K}^{g}}
\newcommand{\bKtheta}{\textbf{K}^{\Gamma}}
\newcommand{\bmu}{\boldsymbol{\mu}}

\newcommand{\boldf}{\textbf{f}}
\newcommand{\bg}{\textbf{g}}

\newcommand{\Ystar}{\Tilde{\mathcal{Y}}}
\newcommand{\ystar}{\Tilde{y}}
\newcommand{\Ycens}{\mathcal{Y}}

\usepackage{caption} 
\title{Generalized Multi-Output Gaussian Process Censored Regression}

\author{%
  Daniele Gammelli \\
  Technical University of Denmark\\
  \texttt{daga@dtu.dk} \\
  \And
  Kasper Pryds Rolsted \\
  Technical University of Denmark\\
  \texttt{kasper@pryds-rolsted.dk} \\
  \And
  Dario Pacino \\
  Technical University of Denmark\\
  \texttt{darpa@dtu.dk} \\
   \And
  Filipe Rodrigues \\
  Technical University of Denmark\\
  \texttt{rodr@dtu.dk} \\
}

\begin{document}

\maketitle

\setcounter{page}{20}

\begin{abstract}
When modelling censored observations (i.e. data in which the value of a measurement or observation is un-observable beyond a given threshold), a typical approach in current regression methods is to use a censored-Gaussian (i.e. Tobit) model to describe the conditional output distribution.
In this paper, as in the case of missing data, we argue that exploiting correlations between multiple outputs can enable models to better address the bias introduced by censored data.
To do so, we introduce a heteroscedastic multi-output Gaussian process model which combines the  non-parametric flexibility of GPs with the ability to leverage information from correlated outputs under input-dependent noise conditions.
To address the resulting inference intractability, we further devise a variational bound to the marginal log-likelihood suitable for stochastic optimization.
We empirically evaluate our model against other generative models for censored data on both synthetic and real world tasks and further show how it can be generalized to deal with arbitrary likelihood functions.
Results show how the added flexibility allows our model to better estimate the underlying non-censored (\mbox{i.e.} true) process under potentially complex censoring dynamics.
\end{abstract}

\section{Introduction}
\label{sec:introduction}

Learning well-specified probabilistic models capable of dealing with censored data is a long-standing challenge of the statistical sciences and machine learning.
Censoring represents the scenario in which the value of a given observation or measurement is only partially known - e.g. survival time in clinical trials, value of interest occurring outside the range of a measuring instrument, observable demand upper-bounded by available supply, etc.
Because of the generality of its definition, censoring arises in numerous domains and is thus historically relevant in multiple research fields \cite{RanganathEtAl2016, JingEtAl2018}.
Most importantly, from a learning perspective, if the dependent variable is censored for a non-neglectable fraction of the observations, parameter estimates obtained by standard regression approaches (e.g. OLS) are inherently biased.   

\smallskip Amongst classical statistical approaches, Tobit models \cite{Tobin1958} have characterized a unifying probabilistic framework capable of accounting for censored data through its likelihood function.
Despite their attractive probabilistic interpretation, Tobit models have typically been limited to relatively simple parameterizations (e.g. linear models) \cite{AllikEtAl2016, Terza1985}.
More recently, evidence has been gathered in favor of combinations bringing together the capabilities of censored models with more flexible architectures such as deep neural networks \cite{WuEtAl2018} and non-parametric models such as Gaussian processes \cite{GammelliEtAl2020, GrootEtAl2012} and random forests \cite{LiEtAl2020}. 

\smallskip From a probabilistic standpoint, censored regression can be framed as the problem of recovering an unobservable, i.e. latent, function by only having access to its observable censored realization.
In light of this, a specially appealing framework for censored regression is provided by Gaussian processes (GP).
By explicitly expressing a distribution over functions, we argue that GPs provide a natural framework for dealing with censored data and inferring the latent non-censored process from censored observations.
Of particular interest to this work are multi-output Gaussian processes (MOGP), which extend the flexible GP framework to the vector-valued random field setup \cite{AlvarezEtAl2012}, showing how it is possible to obtain better predictive performance by exploiting correlations between multiple outputs across the input space, especially in situations affected by missing or noisy data \cite{BonillaEtAl2008}.
In this paper, just as for the case of missing data, we suggest that exploiting correlations between multiple outputs can enable models to acquire a better understanding of the underlying non-censored process.
Crucially, we argue that by jointly modeling multiple outputs, censored models will not only benefit from the correlation between the observable processes, but, most importantly, also from correlations in the censoring process (i.e., the process determining which observations are affected by censoring).
We also draw the connection between heteroscedastic regression and censored modelling, showing how the assumption of input-dependent noise can enable for better modelling of censored data in the context of Tobit likelihoods.
In this paper, we propose a heteroscedastic multi-output censored Gaussian process model (HMOCGP) as a general approach to deal with vector-valued censored data. 
We further show how the proposed model can easily be extended to deal with arbitrary non-Gaussian likelihood functions and the potential impact of inappropriate assumptions regarding the distribution of the residuals in terms of predictive performance. 
We evaluate the proposed HMOCGP against various state-of-the-art GP-based approaches for censored data modelling in the context of an ablation study on both synthetic and real-world tasks. 
In particular, we focus on the problem of recovering the true, \mbox{i.e.} non-censored, function, to which we only have partial access through censored observations. 
For the explored tasks, we show how the ability to leverage information from correlated sources together with the possibility of modelling input-dependent noise, allows HMOCGPs to outperform the baselines in modelling the underlying non-censored signal.

To summarize, the main contributions of this paper are the following:
\begin{itemize}
    \item we propose a novel extension to the multi-output Gaussian process framework that leverages information from multiple correlated outputs in order to address the censoring problem;
    
    \item we study the importance of heteroscedastic approaches and how these can achieve better predictive performance in the context of censored data;
    
    \item we position the proposed model into a general framework capable of dealing with arbitrary likelihood functions for the purpose of censored modelling;  
    
    \item we evaluate the proposed model through an ablation study and compare it with several state-of-the-art approaches on both synthetic and real world datasets.
\end{itemize}

\section{Related Work}
\label{sec:related_work}

\subsection{Censored Modeling}
Historically, learning well-specified models of censored data has always attracted a lot attention in statistical research.
Since their introduction, Tobit models have characterized a unifying probabilistic approach proving that, if the dependent variable in a regression model is censored for a significant fraction of the observations, the method of least squares estimation is inappropriate.
However, in the context of censored models, inference is no longer analytically tractable, thus approximations become necessary during learning.
For this reason, over the years, a number of estimation procedures have been developed.
For example, in \cite{Powell1986}, the author extends the Least Absolute Deviation (LAD) estimator to more general quantiles than the median, in order to better characterize the conditional output distribution.

On the other hand, from a Bayesian perspective, \cite{Chib1992} adopts Monte Carlo integration and Laplace's approximation for the task of posterior inference in the context of Tobit models.
However, classical approaches to censored modelling have usually been characterized by simple parameterizations, such as linear models. 
Related to our work, in \cite{GrootEtAl2012}, the authors also extend the framework of Gaussian processes to the censored data regime, in which they develop an Expectation Propagation algorithm for posterior inference and where they show how the non-parametric nature of GPs can bring to better predictive performance on synthetic datasets.
In our work, we build on these advances in a number of different ways, namely: (i) we assume the censoring threshold to be non-constant and itself a function of $\bx$, thus enabling our proposed model to be effectively used in real world applications where e.g., the supply typically varies in time, (ii) we assume non-constant heteroscedastic observation noise and account for different likelihood functions, (iii) we use multi-output Gaussian process priors to model the latent parameters, thus allowing our proposed model to exploit information from potentially correlated outputs and (iv) we devise a variational bound to the marginal log-likelihood which we optimize through stochastic variational inference for scalable posterior approximation.

\subsection{Gaussian Processes}
There exists an extremely vast literature regarding Gaussian processes, with research focusing on both theoretical and applied contributions.
Within this line of literature, our work is strongly tied to a number of research directions, which we categorize as follows: (i) multi-output extensions of GPs, (ii) heteroscedastic GPs, (iii) generalization of GPs to the case of arbitrary likelihood functions, and (iv) variational methods for GPs (for a more general overview of the field, we refer the reader to \cite{RasmussenWilliams2006}).

\textit{Multi-output GPs:} \hspace{1mm} Largely pioneered in the field of geostatistics, the use of Gaussian processes for the estimation of multi-output signals is also known as \emph{cokringing}.
Within this line of research, a prominent approach to define valid multivariate models is represented by the linear model of coregionalization (LMC) \cite{Goovaerts1997, Journel1976}.
At a high level, probabilistic models based on the LMC achieve multi-output learning by defining the covariance function as the sum of the products of two covariance functions, one that models the dependence between the outputs (i.e. inter-task dependency), and one that models the input dependence (i.e., intra-task dependecy).
Crucially, the majority of current multi-task kernel-based methods can be framed in the context of the LMC, where a notable example is the intrinsic coregionalization model (ICM) \cite{Goovaerts1997}.
In this context, Hierarchical Gaussian processes\cite{LiEtAl2018} have also been used to approach the multi-task learning problem.

\textit{Heteroscedastic GPs:} \hspace{1mm} Standard Gaussian processes assume observation noise to be constant throughout the entire input space. 
In practice however, this assumption does not always hold.
In \cite{LazaroEtAl2020}, the authors introduce a variational approximation allowing for efficient inference in heteroscedastic GP models, whereby input-dependent noise is modelled by placing a GP prior over the variance parameter of a Gaussian noise term.
A similar approach is also introduced in \cite{MunozEtAl2018}, where the input-dependent noise assumption is placed in the context of a multi-output formulation and in \cite{PlataniosEtAl2014}, where noise is modeled as a latent model component switching procedure through a non-parametric Bayesian mixture model.

\textit{Generalized GPs:} \hspace{1mm} Another relevant line of literature for this work is the one focusing on the extension of standard Gaussian processes to arbitrary likelihood functions.
In \cite{SaulEtAl2016}, by developing a variational approximation for the proposed Chained Gaussian Processes framework, the authors propose a viable formulation to train GP-based models for potentially non-Gaussian observations.
In a related way, in \cite{MunozEtAl2018} the authors also propose a formulation to extend standard MOGPs to deal with combinations of arbitrary likelihoods.
Another example of this idea can be found in the Log Gaussian Cox Process (LGCP) model, whereby observation are assumed to be generated by a conditional Poisson distribution with (log) rate defined by a Gaussian process \cite{Moller1998}.

\textit{Variational approximations for GPs:} 
Within the GP literature, approximate inference techniques have attracted much interest in the context of (i) non-conjugate GP models, and (ii) reducing the computational cost of inference in GPs (typically $\mathcal{O}(N^3)$, where N is the number of data points). 
In scenarios where the likelihood is non-Gaussian, the posterior and marginal likelihood must be approximated.
In \cite{NickischEtAl2008}, the authors provide a structured comparison of several approximation techniques for the case of (binary) GP classification, such as Laplace approximations, Markov Chain Monte Carlo methods and Variational bounds.
On the other hand, to reduce computational complexity, several works focus on low-rank or \emph{sparse} approaches, typically relying on the use of \emph{inducing points}. 
However, selecting the location of inducing points can result in a difficult optimization problem.
\cite{Titsias2009} introduces a variational formulation for sparse approximations allowing for the joint estimation of the inducing points and kernel hyper-parameters.
In \cite{HensmanEtAl2013}, this variational approximation is further extended to be amenable to stochastic optimization, thus allowing GPs to be fitted to large datasets also for non-conjugate GP models \cite{HensmanEtAl2015} and to handle the crowdsourcing problem \cite{RuizEtAl2019, GilGonzalezEtAl2018}.

\subsection{Survival Analysis}
Another line of literature related to this work is the one using GP-based approaches to survival analysis \cite{FernandezEtAl2016, AlaaEtAl2017}, where the presence of censoring represents a distinctive feature of the problem characteristics (e.g. when patients drop out of the study in clinical trials). 
Despite the similar intentions, the use of GPs in our work is fundamentally different.
In particular, a traditional approach in survival analysis would be to define a Gaussian process prior over the hazard function $\lambda$.
On the other hand, in our work, we clearly differ from this body of literature by assuming our data to be generated by a specific parametric family of distributions (e.g. censored-Gaussian, censored-Poisson, etc.), and by defining a multi-output Gaussian process prior over the parameters of the given distribution (e.g. mean and variance of a censored-Gaussian distribution).  

\section{Heteroscedastic MOCGP Model}
\label{sec:hmocgp}

\noindent Consider a set of $D$ output functions, $\Ystar = \{\ystar_d(\bx)\}_{d=1}^{^D}$, which we want to jointly model using Gaussian processes.
Traditionally, the literature has considered the case in which each $\ystar_d(\bx)$ is fully observable (i.e. non-censored), continuous and Gaussian distributed (although approaches have extended MOGPs to deal with a mix of continuous, categorical, binary or discrete variables \cite{MunozEtAl2018}). 
In this work, we are interested in the censored case in which we do not have full observability over the set of outputs $\Ystar$, but rather only have access to its censored version $\Ycens = \{y_d(\bx)\}_{d=1}^{^D}$.
In particular, without loss of generality, we will assume to be dealing with the case of right-censored data, such that, for each output $d = 1,\ldots,D$, the censored function $y_d(\bx)$ is upper-bounded by a given threshold $y_d^u(\bx)$ as follows:
\begin{equation}
y_d(\bx)=\begin{cases}
  \ystar_d(\bx), & \text{if $\ystar_d(\bx) < y_d^u(\bx)$} \\
  y_d^u(\bx), & \text{if $\ystar_d(\bx) \geq y_d^u(\bx)$}.
\end{cases}
\label{eq:censoring}
\end{equation}

\smallskip \noindent We note that the threshold $y_d^u(\bx)$ is non-constant and can itself be a function of $\bx$.

Traditionally, when modelling scalar-valued censored data, a reasonable choice of an observation model is a Tobit likelihood. In this context, the probability of $y$ under the censored-Gaussian distribution is:
\begin{align}
p(y | \bx) & =\mathcal{N}(y | \mu(\bx), \sigma^2)^{\mathbbm{1}_{\bx \notin \mathcal{C}}} \label{eq:tobit}\\
& \hspace{4mm}( 1 - \Phi(y | \mu(\bx), \sigma^2))^{\mathbbm{1}_{\bx \in \mathcal{C}}}, \nonumber
\end{align}
\noindent where $\mathbbm{1}$ is the indicator function with $\mathcal{C}$ defined as the set of censored observations (i.e., $\bx \notin \mathcal{C}$ if $y < y_d^u(\bx)$ and $\bx \in \mathcal{C}$ otherwise) and where $\Phi$ is the Gaussian cumulative density function (CDF). Typically, $\mu(\bx)$ is assumed to be a (linear) parametric function (e.g. $\mu(\bx) = \mathbb{\beta}^\top \bx$), whose parameters are estimated, together with the variance $\sigma^2$, through e.g. maximum likelihood. This likelihood is well known in the literature as a Tobit likelihood \cite{Demaris2005, Greene2012}, or a type I Tobit model according to the taxonomy of \cite{Amemiya1984}. Intuitively, the Tobit likelihood lends itself to a natural interpretation of the censoring problem by (i) incentivizing parameter estimates to assign high probability to those observations for which we have access to the true value (through the PDF term in Eq. \ref{eq:tobit}), and (ii) using the $1 - \Phi(\cdot)$ term to measure the probability of a censored observation being greater or equal than the observed value $y$ (thus, naturally encoding the concept of right-censored data). 

In this section, we extend the above concepts to the vector-valued random field setup, and define the generative model $p_{\theta}$ and variational family $q_{\phi}$ characterizing the proposed HMOCGP for the purpose of censored data modelling.
The resulting intractability of the posterior distribution is further approached by learning a tractable approximation through \emph{stochastic variational inference} (SVI) \cite{HoffmanEtAl2013, BleiEtAl2017}.

\subsection{Generative model} 

Given the set of censored observations $\mathcal{C}$ (i.e., we know which input vectors $\bx \in \bX$ are affected by censoring), we will assume each $y_d(\bx)$ to be distributed according to a censored-Gaussian defined by mean $f_d(\bx)$ and variance $g_d(\bx)$. 
To ensure positivity, we parameterize $ \eta({g_d(\bx)})$, where $\eta(\cdot)$ is any deterministic function that maps $g_d(\cdot)$ to the set of positive real numbers (e.g. $\eta(\cdot)$ could be modelled using an exponential or a softplus function).
Concretely, let us define a vector-valued function $\by(\bx) = [y_1(\bx), \ldots, y_D(\bx)]^\top$. We assume that the outputs are conditionally independent given the latent function values $\boldf(\bx) = [f_1(\bx), \ldots, f_D(\bx)]^\top$ and $\bg(\bx) = [g_1(\bx), \ldots, g_D(\bx)]^\top$ such that:
\begin{align}
p(\by(\bx) | \boldf(\bx), \bg(\bx)) & = \prod_{d=1}^D p(y_d(\bx) | f_d(\bx), \eta(g_d(\bx))) = \nonumber\\
& = \prod_{d=1}^D \mathcal{N}(y_d(\bx) | f_d(\bx), \eta(g_d(\bx)))^{\mathbbm{1}_{\bx \notin \mathcal{C}}} \label{eq:likelihood}\\
& \hspace{4mm} ( 1 - \Phi(y_d(\bx) | f_d(\bx), \eta(g_d(\bx))))^{\mathbbm{1}_{\bx \in \mathcal{C}}}. \nonumber
\end{align}


\smallskip Another of our main departures from previous work in the context of censored modelling is in placing multi-output Gaussian process priors on $\boldf(\bx)$ and $\bg(\bx)$ that allow for correlations between censored functions under input-dependent noise conditions. Let us define $\Gamma = \{f, g\}$ to jointly refer to both latent mean and variance parameters of the censored-Gaussian distribution, respectively (e.g. $\Gamma_d(\bx) = \{f_d(\bx), g_d(\bx)\}$).
We use a covariance function based on the linear model of coregionalization (LMC) to express correlations between functions $\Gamma_d(\bx)$ and $\Gamma_{d'}(\bx)$.
Concretely, consider an additional set of independent latent functions $\mathcal{U} = \{u_q(\bx)\}_{q=1}^Q$ that will be linearly combined to produce $D$ latent vectors $\{\Gamma_d(\bx)\}_{d=1}^D$.
The latent functions $u_q(\bx)$ are then assumed to be drawn independently from $Q$ independent GP priors such that $u_q(\cdot) \sim \mathcal{GP}(\mathbf{0}, k_q(\cdot, \cdot))$, where $k_q$ can be defined as any valid covariance function, or kernel, and where for simplicity we assume zero mean. 
Each latent parameter is then given by:
\begin{equation}
\Gamma_d(\bx) = \sum_{q=1}^{Q} \sum_{i=1}^{R_q} a_{d, q}^i u_q^i(\bx), 
\label{eq:lmc}
\end{equation}

\noindent where $u_q^i$ are $R_q$ i.i.d. samples from $u_q(\cdot) \sim \mathcal{GP}(\mathbf{0}, k_q(\cdot, \cdot))$ and $a_{d, q}^i \in \mathbb{R}$ are learnable parameters controlling the linear combination.
Given this definition, the mean function for $\Gamma_d(\bx)$ is zero and the cross-covariance function $k_{\Gamma_d, \Gamma_{d'}} = \text{cov}[\Gamma_d(\bx), \Gamma_{d'}(\bx)]$ is equal to $\sum_{q=1}^{Q} b_{d, d'}^q k_q(\bx, \bx')$, where $b_{d, d'}^q = \sum_{i=1}^{R_q} a_{d, q}^i a_{d', q}^i$.

\smallskip Let $\bX = \{ \bx_n\}_{n=1}^{N}$ be the set of common input vectors for all outputs $y_d(\bx)$. 
The generative process for the heteroscedastic MOCGP is as follows. 
We sample $\boldf \sim \mathcal{N}(\mathbf{0}, \bKmu)$ and $\bg \sim \mathcal{N}(\mathbf{0}, \bKsigma)$, where both $\bKmu$ and $\bKsigma$ are block-wise matrices with blocks given by $\{\bKmu_{f_d, f_{d'}}\}_{d=1, d'=1}^{D, D}$ and $\{\bKsigma_{g_d, g_{d'}}\}_{d=1, d'=1}^{D, D}$, respectively.
For simplicity, we will again use $\Gamma = \{f, g\}$ to jointly refer to both mean and variance functions (e.g., $\bK^{\Gamma} = \{\bK^f, \bK^g\}$).
The elements in $\bKtheta$ are then given by applying the kernel function $k_{\Gamma_d, \Gamma_{d'}}(\bx_n, \bx_m)$, between any two input vectors $\bx_n, \bx_m \in \bX$.
In the case of equal inputs $\bX$ for all latent parameter vectors, $\bKtheta$ can be expressed as the sum of Kroenecker products 
\begin{equation}
\bKtheta = \sum_{q=1}^{Q} \bA_q^\Gamma {\bA_q^\Gamma}^\top \otimes \bKtheta_q = \sum_{q=1}^{Q} \bB_q^\Gamma \otimes \bKtheta_q,
\label{eq:lmc_covar}
\end{equation}
where $\bA_q^\Gamma \in \mathbb{R}^{D \times R_q}$ and $\bB_q^\Gamma \in \mathbb{R}^{D \times D}$ have elements $\{a_{d, q}^{\Gamma, i}\}_{d=1, i=1}^{D, R_q}$ and $\{b_{d, d'}^{\Gamma, q}\}_{d=1, d'=1}^{D, D}$ respectively and where $\bKtheta_q \in \mathbb{R}^{N \times N}$ has entries defined by $k_q^\Gamma(\bx_n, \bx_m)$ for $\bx_n, \bx_m \in \bX$.
In literature, the matrices $\bB_q^\Gamma$ are known are \emph{coregionalization matrices}.
Once we obtain the samples $\boldf$ and $\bg$, we can generate the elements in $\by$ by sampling from the censored-Gaussian conditional distribution $p(\by(\bx) | \boldf(\bx), \bg(\bx))$ as defined in Eq. \ref{eq:likelihood}. 

In practice, the proposed model can be summarized by the following generative process:

\begin{algorithmic}[1]
    \STATE Given a dataset $\bX = \{ \bx_n\}_{n=1}^{N}$, kernel functions $k_q^{\Gamma}(\cdot, \cdot) = \{k_q^{f}(\cdot, \cdot), k_q^{g}(\cdot, \cdot)\}$, matrices $\bB_q^{\Gamma} = \{\bB_q^{f}, \bB_q^{g}\}$, with $q = 1, \ldots, Q$ and the set of censored observations $\mathcal{C}$ 
  \STATE Compute covariance matrices $\bK^\Gamma = \{\bK^f, \bK^g\}$ as in Eq.\ref{eq:lmc_covar}
  \STATE Sample $\boldf \sim \mathcal{N}(\mathbf{0}, \bKmu)$
  \STATE Sample $\bg \sim \mathcal{N}(\mathbf{0}, \bKsigma)$
  \FOR{$\bx_n$ in $\bX$}
    \STATE Sample $\by(\bx_n) \sim p(\by(\bx_n) | \boldf(\bx_n), \bg(\bx_n))$ as defined in Eq. \ref{eq:likelihood}.
  \ENDFOR
\end{algorithmic}

\subsection{Inference}
Given the non-Gaussian likelihood in the definition of the HMOCGP, exact posterior inference is intractable. 
Hence, given a dataset $\mathcal{D} = \{\bX, \by\}$, we use variational inference (VI) to jointly optimize the model's hyper-parameters and compute a tractable approximation to the posterior distribution over $\boldf$ and $\bg$.
Through VI, we use ideas from the calculus of variations to find a parametric approximation $q_{\phi}(\mathbf{f}_{d},\mathbf{g}_{d})$ that minimizes a measure of dissimilarity with respect to the true, intractable posterior.
By doing so, variational methods allow us to reduce a complex inference problem into a simpler optimization problem (for a more in-depth review of variational methods, we refer the reader to \cite{BleiEtAl2017}).

Denoting $\theta$ and $\phi$ as the set of model and variational
parameters respectively, VI offers a scheme for jointly optimizing parameters $\theta$ and computing an approximation to the posterior distribution (through $\phi$) by maximizing the following \emph{evidence lower bound} (i.e. ELBO), where we omit the dependency on $\bX, \by$ for simplicity:
\begin{align}
    \log p_{\theta}(\mathbf{y}) & = \log \int p_{\theta}(\mathbf{y}, \mathbf{f}, \mathbf{g}) \,  d\mathbf{f} \, d\mathbf{g} \nonumber\\
    & = \log \int \frac{q_{\phi}(\mathbf{f}, \mathbf{g})}{q_{\phi}(\mathbf{f}, \mathbf{g})} p_{\theta}(\mathbf{y}, \mathbf{f}, \mathbf{g}) \, d\mathbf{f} \, d\mathbf{g} \nonumber\\
    & = \log \mathbb{E}_{q_{\phi}(\mathbf{f}, \mathbf{g})} \left[\prod_{d=1}^{D} \frac{p_{\theta}(\mathbf{y}_d | \mathbf{f}_d, \mathbf{g}_d) p_{\theta}(\mathbf{f}_d) p_{\theta}(\mathbf{g}_d)}{q_{\phi}(\mathbf{f}_d, \mathbf{g}_d)} \right] \nonumber\\
    & \geq \mathbb{E}_{q_{\phi}(\mathbf{f}, \mathbf{g})} \left[\sum_{d=1}^{D} \log p_{\theta}(\mathbf{y}_d | \mathbf{f}_d, \mathbf{g}_d)\right] + \label{eq:elbo}\\ 
    & - \sum_{d=1}^{D} \mathbb{KL} \left(q_{\phi}(\mathbf{f}_{d},\mathbf{g}_{d})|| p_{\theta}(\mathbf{f}_{d}) p_{\theta}(\mathbf{g}_{d}) \right) = \mathcal{L}(\theta, \phi), \nonumber
\end{align}
\noindent where $\theta$ is the set of model hyper-parameters - such as the coregionalization matrices $\bB_q^\Gamma$ and other kernel-specific hyper-parameters (e.g. the lengthscale in an RBF kernel) - and where $\phi$ is the set of variational parameters characterizing the approximate posterior distribution.
According to the theory of VI \cite{BleiEtAl2017}\, maximizing the ELBO in Eq. \ref{eq:elbo} is equivalent to minimizing the Kullback-Leibler (KL) divergence between the variational approximation $q_{\phi}$ and the true posterior distribution of the latent variables in the model.
Resembling a mean-field approximation, we define the following factorization for the variational distribution:
\begin{align}
    q_{\phi}(\boldf, \bg) & = \prod_{d=1}^{D}q_{\phi}(\boldf_d) \, q_{\phi}(\bg_d) \label{eq:q}\\
    & = \prod_{d=1}^{D} \mathcal{N}(\boldf_d | \bmu_{f_d}, \bS_{f_d}) \, \mathcal{N}(\bg_d | \bmu_{g_d}, \bS_{g_d}), \nonumber
\end{align}

\noindent where the variational parameters $\phi = \{\bmu_{f_d}, \bmu_{g_d}, \bS_{f_d}, \bS_{g_d} \}_{d=1}^D$ must be optimized.
In our implementation, we represent each covariance matrix through its lower-triangular Cholesky factorization $\bS_{\{f_d, g_d\}} = \bL \bL^\top$ and estimate $\bL$ instead of $\bS$ to ensure positive definiteness.

\subsection{Predictive Distribution}
The predictive distribution for a test input $\bx_*$ can be approximated as 
\begin{align}
    p(\by_* | \, \by) & = \int p_{\theta}(\by_* \, | \, \boldf_*, \, \bg_*) \, p_{\theta}(\boldf_*, \, \bg_* \, | \, \by) \, d\boldf_* \, d\bg_*\nonumber\\
    & \approx \int p_{\theta}(\by_* \, | \, \boldf_*, \, \bg_*) \, q_{\phi}(\boldf_*, \, \bg_*) \, d\boldf_* \, d\bg_* \label{eq:pred}\\ 
    & = \int \prod_{d=1}^D p_{\theta}(y_d | f_{*,d}, \, g_{*,d}) \,  q_{\phi}(f_{*,d}, \, g_{*,d}) df_{*,d} \, dg_{*,d} , \nonumber
\end{align} 
\noindent where $q_{\phi}(f_{*,d}, \, g_{*,d}) = q_{\phi}(f_{*,d}) \,  q_{\phi}(g_{*,d})$, and where $\boldf_* = f(\bx_*)$, $\bg_* = g(\bx_*)$.
If we then consider $q_{\phi}^*(f_{d}, g_{d})$ as the variational approximation optimized through SVI, $q_{\phi}(f_{*,d})$ and $q_{\phi}(g_{*,d})$ can be obtained as follows:
\begin{align}
    q_{\phi}(f_{*,d}) = \int p_{\theta}(f_{*, d} | \, \bx_{*,d}, \, \boldf_{d}) \, q_{\phi}^*(\boldf_d) \, d\boldf_d \\
    q_{\phi}(g_{*,d}) = \int p_{\theta}(g_{*, d} | \, \bx_{*,d}, \, \bg_{d}) \, q_{\phi}^*(\bg_d) \,  d\bg_d,
\end{align}

\noindent which requires evaluating the model's kernel function at $\bx_*$. In practice, the likelihood term $p_{\theta}(\by_* | \, \boldf_*, \, \bg_*)$ makes the integral in Eq. \ref{eq:pred} intractable, hence, as in the case of the evidence lower bound, we resort to Monte Carlo methods to obtain an approximation.
Amongst other possible approaches to solve the above predictive integral, we choose Monte Carlo methods because of their proven scalability with respect to the problem's dimensionality (opposed to alternative methods such as e.g. Gaussian Hermite quadrature).

\subsection{Generalization to Arbitrary Likelihood Functions}
So far we have considered the case in which each output $y_d(\bx)$ is continuous and censored-Gaussian distributed.
As in \cite{MunozEtAl2018}, where the authors deal with a mix heterogeneous outputs, let us now extend the concepts introduced in previous sections to deal with arbitrary censored distributions.
In particular, we will assume the distribution over outputs $\by(\bx)$ to be completely specified by a set of parameters $\boldsymbol{\gamma}(\bx) \in \mathcal{X}^{D \times J}$, where $\mathcal{X}$ is a generic domain for the parameters, $D$ is again the number of output functions considered, and $J$ is the number of parameters that define the distribution for each output $\by_d(\bx)$.
As in the censored-Gaussian case, we assume the latent parameter vector $\boldsymbol{\gamma}_d(\bx)$ to be specified as a non-linear differentiable transformation of multi-output Gaussian process priors such that $\boldsymbol{\gamma}_d(\bx) = [ \eta_1(\gamma_{d, 1}(\bx)), \ldots,  \eta_J(\gamma_{d, J}(\bx))]$, where $\eta_j$ is again a deterministic function mapping the GP output $\gamma_{d,j}$ to the appropriate domain.

\smallskip To make the notation concrete, let us assume $D=2$ correlated count variables taking values $y_d(\bx) \in \mathbb{N} \cup \{0\}$ that we wish to model using a Poisson distribution specified by a single (i.e., $J = 1$) rate parameter for every output $y_d(\bx)$.
Given the positivity constraint of the rate, we can then define $\boldsymbol{\gamma}(\bx) = [\eta_1(\gamma_{1, 1}(\bx)), \eta_1(\gamma_{2, 1}(\bx))]^{\top}$, where we model $\eta_1$ using an exponential function to ensure positive values for the parameter.
Following the same line of reasoning, we can easily use this general notation to redefine the censored-Gaussian model introduced in previous sections by specifying $J=2$ and $\boldsymbol{\gamma}(\bx) = [\eta_1(\gamma_{d, 1}(\bx)), \eta_2(\gamma_{d, 2}(\bx))]^{\top}$, where we could assume $\eta_1$ to be the identity function for the latent mean parameter and $\eta_2$ as an exponential function to ensure positive values for the variance.

\smallskip Let us now define the generative model $p_{\theta}$ in the general context of arbitrary likelihood functions.
Concretely, given a vector-valued function $\by(\bx) = [y_1(\bx), \ldots, y_D(\bx)]^\top$ and a set of parameters $\boldsymbol{\gamma}(\bx) = [\boldsymbol{\gamma}_{1}(\bx), \ldots, \boldsymbol{\gamma}_{D}(\bx)]^{\top}$, Eq. \ref{eq:likelihood} can be re-written as:
\begin{align}
p_{\theta}(\by(\bx) \mid \boldsymbol{\gamma}(\bx)) & = \prod_{d=1}^D p_{\theta}(y_d(\bx) \mid \boldsymbol{\gamma}_d(\bx)) = \nonumber\\
& = \prod_{d=1}^D P(y_d(\bx) \mid \boldsymbol{\gamma}_d(\bx))^{\mathbbm{1}_{\bx \notin \mathcal{C}}} \label{eq:gen_likelihood}\\
& \hspace{4mm} ( 1 - F(y_d(\bx) \mid \boldsymbol{\gamma}_d(\bx)))^{\mathbbm{1}_{\bx \in \mathcal{C}}}, \nonumber
\end{align}

\noindent where $\mathcal{C}$ is again the set of censored observations and where $P$ and $F$ are, respectively, the PDF (or PMF, in the case of discrete random variables) and CDF of the assumed data distribution.

On the other hand, the variational family $q_{\phi}(\boldsymbol{\gamma})$ can be specified as follows:
\begin{align}
    q_{\phi}(\boldsymbol{\gamma}) & = \prod_{d=1}^{D} \prod_{j=1}^{J} q_{\phi}(\boldsymbol{\gamma}_{d, j}) \label{eq:q_gen}\\
    & = \prod_{d=1}^{D} \prod_{j=1}^{J} \mathcal{N}(\boldsymbol{\gamma}_{d, j} \mid \bmu_{\gamma_{d, j}}, \bS_{\gamma_{d, j}}), \nonumber
\end{align}

\noindent where parameters $\phi = \{\bmu_{\gamma_{d, j}}, \bS_{\gamma_{d, j}} \}_{d=1, j=1}^{D, J}$ must be jointly optimized with kernel hyper-parameters $\theta$ by maximizing the following variational lower bound:
\begin{align}
    \log p_{\theta}(\mathbf{y})
    & \geq \mathbb{E}_{q_{\phi}(\boldsymbol{\gamma})} \left[\sum_{d=1}^{D} \log p_{\theta}(\mathbf{y}_d | \boldsymbol{\gamma}_{d})\right] + \label{eq:gen_elbo}\\ 
    & - \sum_{d=1}^{D}\sum_{j=1}^{J} \mathbb{KL} \left(q_{\phi}(\boldsymbol{\gamma}_{d, j})|| p_{\theta}(\boldsymbol{\gamma}_{d, j}) \right) = \mathcal{L}(\theta, \phi). \nonumber
\end{align}

\noindent Finally, the predictive distribution can be approximated by re-defining Eq. \ref{eq:pred} as:
\begin{align}
    p(\by_* | \, \by)
    & \approx \int \prod_{d=1}^D p_{\theta}(y_d | \, \gamma_{*,d}) \,  q_{\phi}(\gamma_{*,d}), \label{eq:gen_pred}
\end{align} 

\noindent where, if we consider $q_{\phi}^*(\boldsymbol{\gamma}_d)$ as the variational approximation optimized through SVI, $q_{\phi}(\gamma_{*,d})$ can be obtained by solving the following integral:
\begin{align}
    q_{\phi}(\gamma_{*,d}) = \int p_{\theta}(\gamma_{*, d} | \, \bx_{*,d}, \, \boldsymbol{\gamma}_{d}) \, q_{\phi}^*(\boldsymbol{\gamma}_d) \, d\boldsymbol{\gamma}_d.
\end{align}

\section{Experiments}
\label{sec:experiments}

In this paper, we are interested in modelling the true, non-censored data distribution. However, we assume to have only partial access to it through a set censored observations. The presence of potentially complex censoring dynamics make this problem particularly relevant from both a methodological and applied standpoint in multiple research fields, like health, demand modelling, physics and many others.
To demonstrate its performance in terms of censored data modelling, we evaluate the proposed HMOCGP on several applications with both synthetic and real-world data.

\subsection{Models}
We evaluate the proposed HMOCGP\footnote{Code available at: \url{https://github.com/DanieleGammelli/multi-output-gp-censored-regression}} against state-of-the-art GP-based approaches to censored data modelling. Concretely, we place our experiments in the context of an ablation study aiming to investigate the effect on performance of the various building blocks which characterize the architectures of the implemented models.
In particular, we compare with the following models:
\begin{itemize}
    \item \emph{Non-Censored Gaussian Process (NCGP)}: a standard single-output GP assuming homoscedastic noise - the most common in literature. 
    It is defined by the following likelihood function:
    \begin{equation*}
        p(y(\bx) \, | \, f(\bx), \, \sigma) = \mathcal{N}(y(\bx) \, | \, f(\bx), \, \sigma^2)
    \end{equation*}
        
    \item \emph{Multi-Output Non-Censored Gaussian Process (MONCGP)} as in \cite{AlvarezEtAl2012}: extends the NCGP by allowing for correlations between multiple outputs as defined by the LMC.
    It is defined by the following likelihood function:
    \begin{equation*}
        p(\by(\bx) \, | \, \boldf(\bx), \, \sigma) = \prod_{d=1}^{D} \mathcal{N}(y_d(\bx) \, | \, f_d(\bx), \, \sigma^2)
    \end{equation*}
    \item \emph{Censored Gaussian Process (CGP)} as in \cite{GammelliEtAl2020, GrootEtAl2012}: a single-output GP assuming censored-Gaussian likelihood.
    It is defined by the following likelihood function:
    \begin{align*}
        p(y(\bx) \, | \, f(\bx), \, \sigma) & = \mathcal{N}(y(\bx) \, | \, f(\bx), \, \sigma^2)^{\mathbbm{1}_{\bx \notin \mathcal{C}}} \\
        & \left(1 - \Phi(y(\bx) \, | \, f(\bx), \, \sigma^2)\right)^{\mathbbm{1}_{\bx \in \mathcal{C}}}
    \end{align*}
    \item \emph{Heteroscedastic Censored Gaussian Process (HCGP)}: extends the CGP by allowing for input-dependent noise.
    It is defined by the following likelihood function:
    \begin{align*}
        p(y(\bx) \, | \, f(\bx), \, g(\bx)) & =  \mathcal{N}(y(\bx) \, | \, f(\bx), \, \eta(g(\bx)))^{\mathbbm{1}_{\bx \notin \mathcal{C}}} \\
        & \left(1 - \Phi(y(\bx) \, | \, f(\bx), \, \eta(g(\bx)))\right)^{\mathbbm{1}_{\bx \in \mathcal{C}}}
    \end{align*}
    \item \emph{Multi-Output Censored Gaussian Process (MOCGP)}: extends the CGP by allowing for correlations between multiple outputs.
    The likelihood function is given by:
    \begin{align*}
        p(\by(\bx) \, | \, \boldf(\bx), \, \sigma) =  & \prod_{d=1}^{D} \mathcal{N}(y_d(\bx) \, | \, f_d(\bx), \, \sigma^2)^{\mathbbm{1}_{\bx \notin \mathcal{C}}} \\
        & \left(1 - \Phi(y_d(\bx) \, | \, f_d(\bx), \, \sigma^2)\right)^{\mathbbm{1}_{\bx \in \mathcal{C}}}
    \end{align*}
\end{itemize}
All models were implemented using PyTorch \cite{PaszkeGrossEtAl2019} and the universal probabilistic programming language Pyro \cite{BinghamEtAl2018}. In our experiments, we use 3 samples to approximate the intractable expectations in the ELBO and its gradients.

\begin{figure*}[t]
\begin{subfigure}{.5\textwidth}
\centering
\includegraphics[width=1\linewidth]{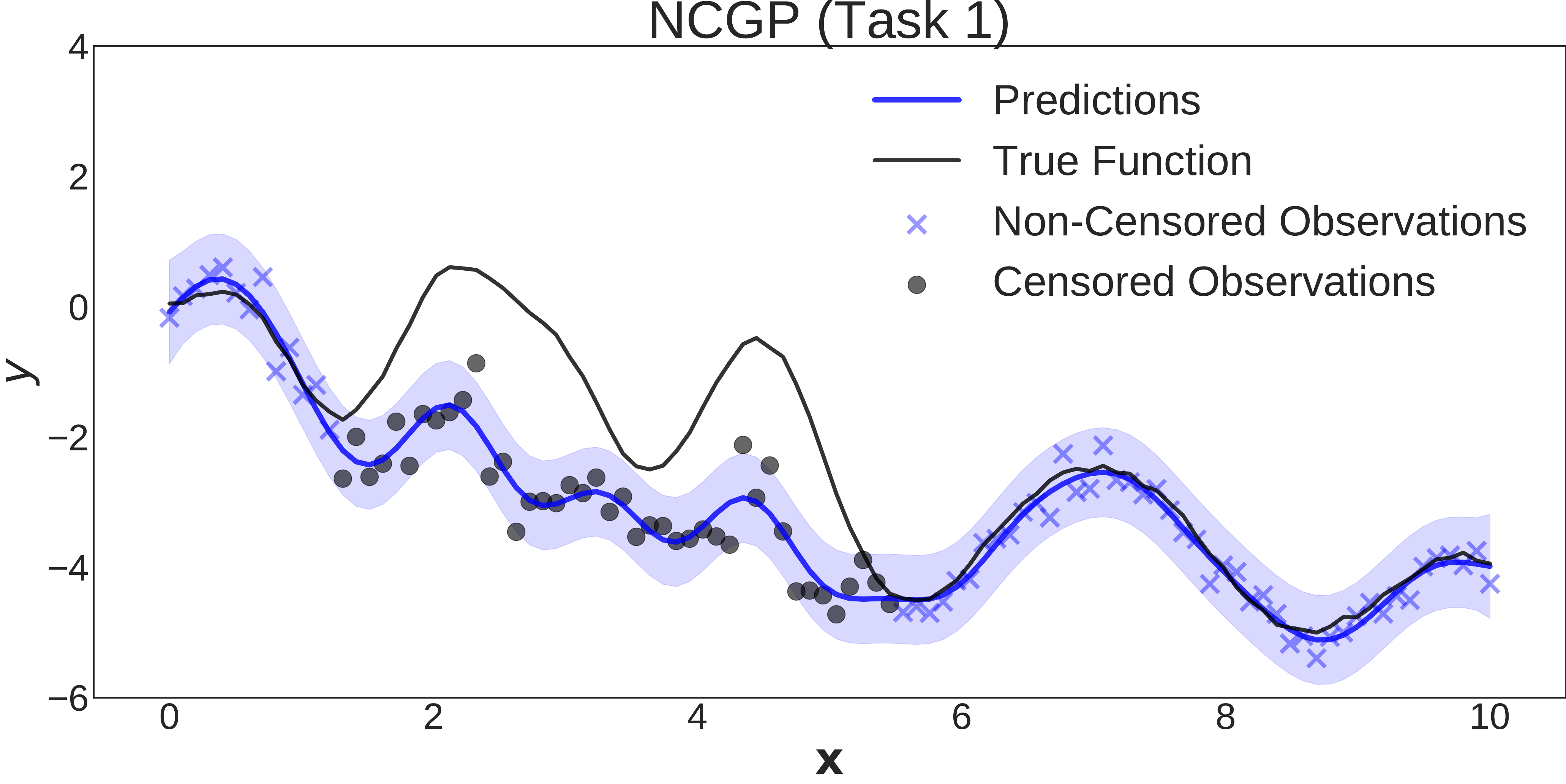}
\end{subfigure}
\hfill
\begin{subfigure}{.5\textwidth}
\centering
\includegraphics[width=1\linewidth]{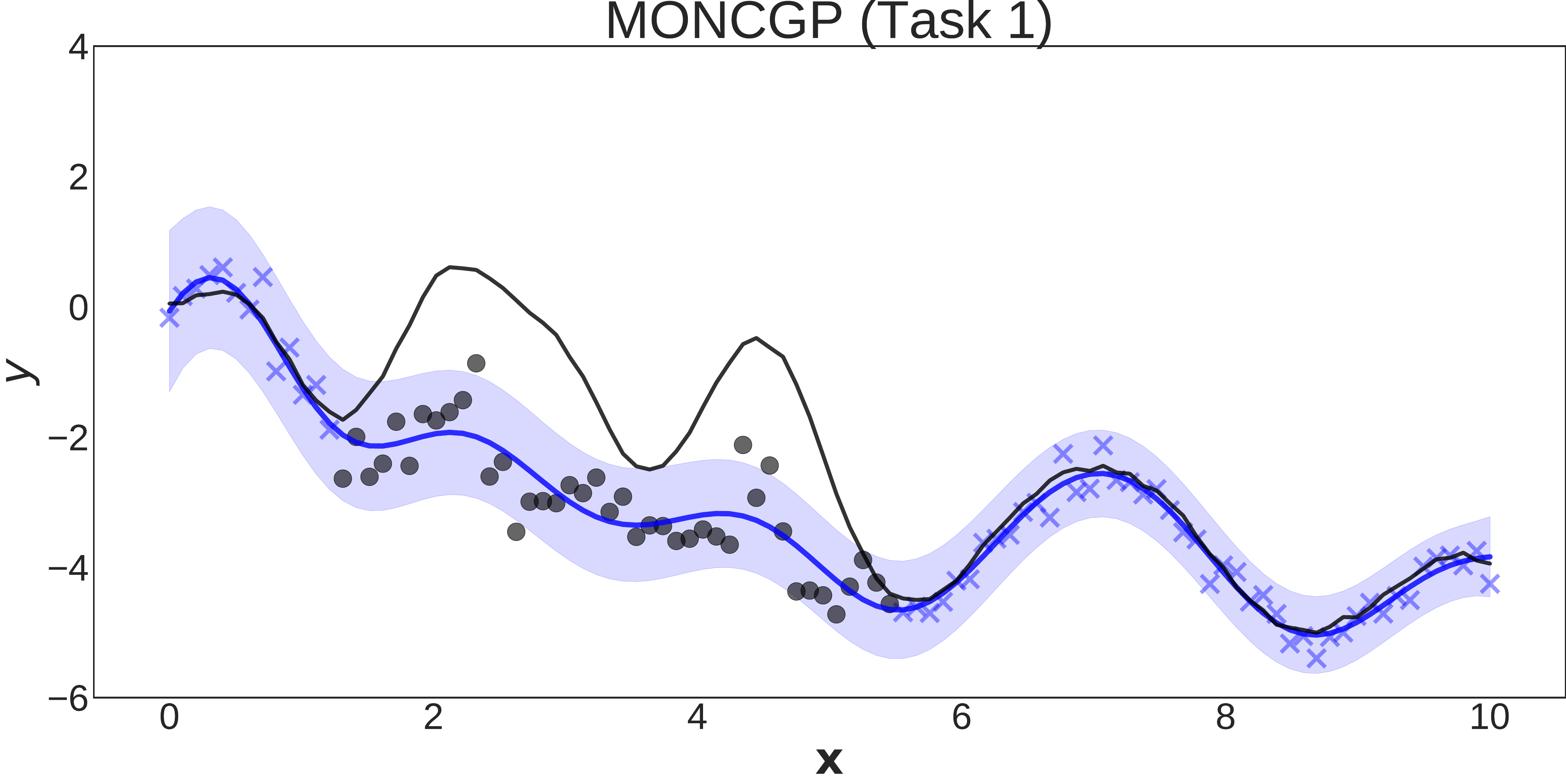}
\end{subfigure}

\begin{subfigure}{.5\textwidth}
\centering
\includegraphics[width=1\linewidth]{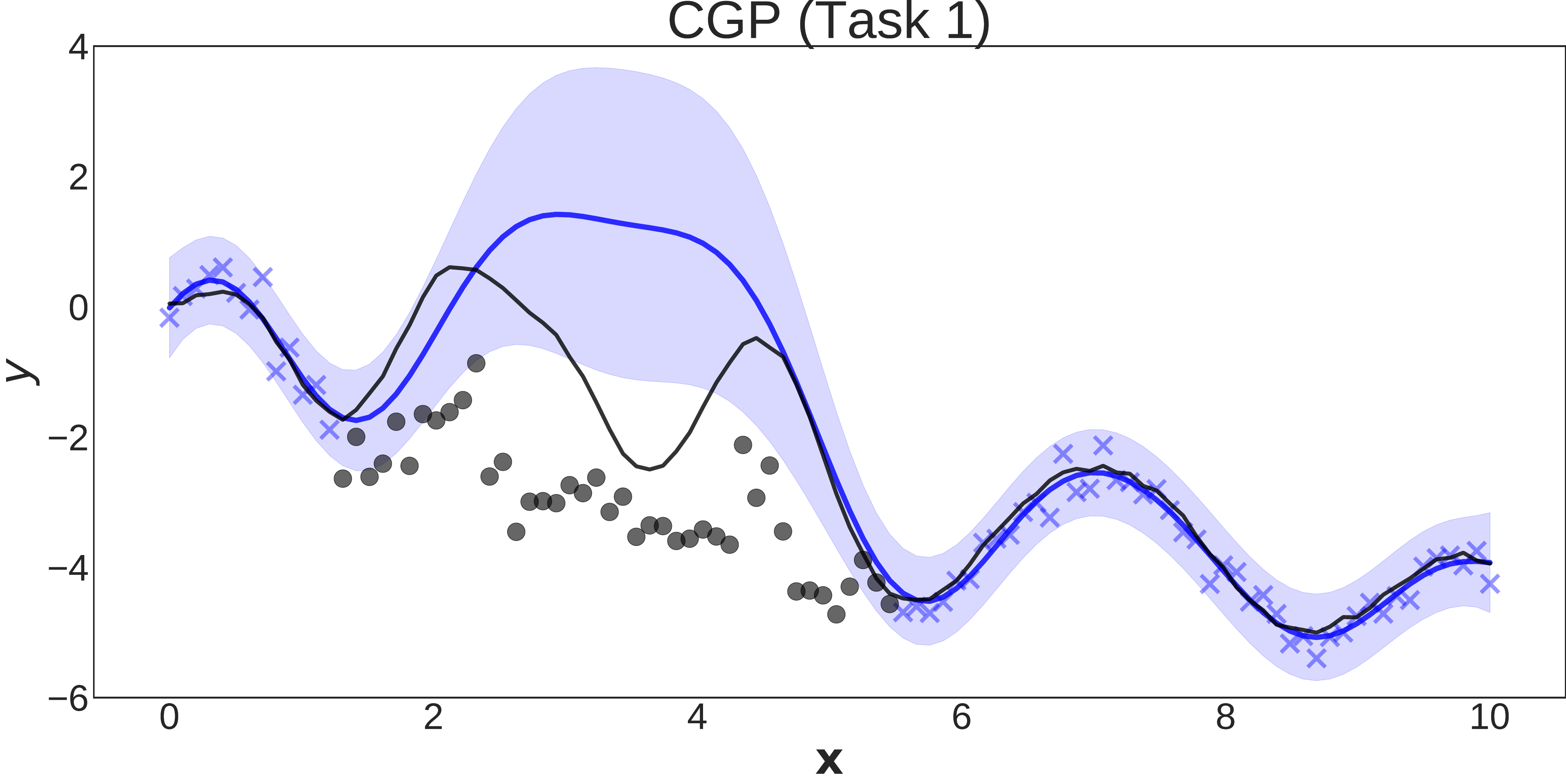}
\end{subfigure}
\begin{subfigure}{.5\textwidth}
\centering
\includegraphics[width=1\linewidth]{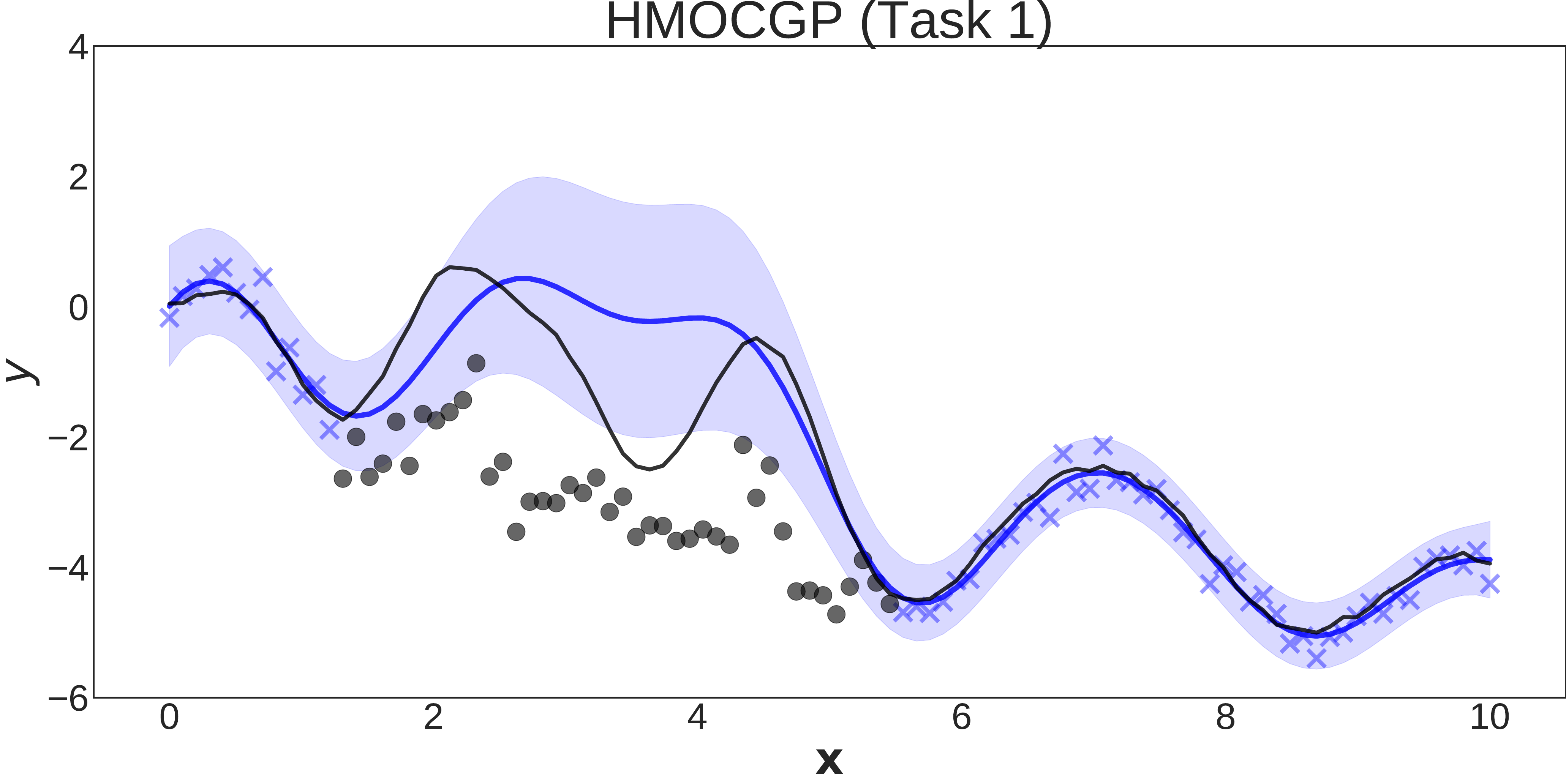}
\end{subfigure}
\caption{Comparison between NCGP (top left), CGP (bottom left), MONCGP (top right) and the proposed HMOCGP (bottom right) for the first output $y_1(\bx)$ on synthetic data. Results show how the Gaussian likelihood assumption biases the NCGP (both single and multi-output) towards the observed data. On the other hand, while both CGP and HMOCGP account for censorship in the estimation of the latent non-censored function, HMOCGP is able to use information from a correlated output to achieve better predictive performance.}
\label{fig:synthetic}
\end{figure*}

\subsection{Training}
We train each model using stochastic gradient ascent on the evidence lower bound $\mathcal{L}(\theta, \phi)$ defined in Eq.~\ref{eq:gen_elbo} using the RMSprop optimizer \cite{TielmanEtAl2012}, with a fixed learning rate of $0.001$.
The final results on real world tasks were selected with an early-stopping procedure based on validation performance. 
The training of the full model (HMOCGP) took around 30 minutes for all real-world datasets considered, using a NVIDIA GeForce RTX 2080 Ti.

\subsection{Synthetic Dataset} 
In our first experiment, we evaluate the extent to which our model can estimate the non-censored latent function by exploiting training information from a correlated output.
We define the problem by sampling $D = 2$ heteroscedastic real-valued outputs $\ystar_1(\bx)$ and $\ystar_2(\bx)$ from a MOGP prior. 
We assume the outputs share a common input set and sample $N_1 = N_2 = 100$ equally spaced samples in the input range $[0, 10]$.
We further assume the first output to be affected by some latent censoring process in the interval $[1.3, 5.5]$ such that $y_1(\bx) \leq \ystar_1(\bx)$. 
On the other hand, we assume no censoring on the second output, thus obtaining $y_2(\bx) = \ystar_2(\bx)$.
Table \ref{tab:synthetic} shows results for the implemented models in terms of negative log-predictive density\footnote{For a formal definition of the NLPD and other performance metrics used in this work, please see \ref{sec:app_performance}} (NLPD) evaluated on the true, non-censored observations.
Qualitatively, in Figure \ref{fig:synthetic} we can see how the ability of learning from correlated outputs enables the model to better reconstruct the latent non-censored process.

To further asses the robustness of HMOCGP to different degrees of censoring severity, we generate a series of scenarios where both outputs are affected by some latent censoring process.
Specifically, we examine four configurations with different degrees of overlap between censored locations for the two outputs (i.e. the amount of input vectors $\bx$ for which both outputs are censored): (i) 0\% overlap, (ii) 50\% overlap, (iii) 100\% overlap with mild interpolation complexity given by discontinuous censoring locations, and (iv) 100\% overlap with hard interpolation complexity given by continuous censoring locations (a visual description of the four tasks is provided in Figure \ref{fig:overlap1} and \ref{fig:overlap2} in the Appendix).

The results in Table \ref{tab:overlap} highlight a number of interesting points.
First, HMOCGP shows an interesting degree of robustness to censoring overlap, even in extreme cases such as full overlap, thus being able to outperform all benchmarks.
Interestingly, in the case where the censored locations (with 100\% overlap) occupy a crucial location in the signal (making interpolation more complex), jointly modeling the tasks seems to be detrimental for the overall performance.

\smallskip
\begin{table}[t]
\centering
\begin{tabular}{p{6cm} c}
    \hline 
     & \textbf{NLPD} \\ [0.2ex] 
    \hline
    Non-Censored GP (NCGP) & 589.82  \\ [0.1ex]
    Multi-Output NCGP (MONCGP) & 374.79 \\[0.1ex]
    Censored GP (CGP) & 88.89  \\ [0.1ex]
    Heteroscedastic CGP (HCGP) & 83.29  \\[0.1ex]
    Multi-Output CGP (MOCGP) & 64.59  \\[0.1ex]
    Heteroscedastic MOCGP (HMOCGP) & \textbf{51.32} \\[0.1ex]
    \hline
    \end{tabular}%
    \caption{Synthetic Dataset Reconstruction Results.}
\label{tab:synthetic}%
\end{table}

\begin{table}[t]
\centering
\begin{tabular}{p{5.9cm} c c c c}
    & \multicolumn{4}{c}{Censoring Overlap} \\
    \hline 
     & 0\% & 50\% & 100\%-M & 100\%-H \\ [0.2ex] 
    \hline
    Non-Censored GP (NCGP) & 589.82 & 256.78 & 256.78 & 589.82 \\ [0.1ex]
    Multi-Output NCGP (MONCGP) & 364.13 & 129.02 & 140.97 & 398.54 \\[0.1ex]
    Censored GP (CGP) & 88.89 & 32.36 & 32.36 & 88.89 \\ [0.1ex]
    Heteroscedastic CGP (HCGP) & 83.29 & 27.92 & 27.92 & \textbf{83.29} \\[0.1ex]
    Multi-Output CGP (MOCGP) & 86.23 & 25.59 & 25.35 & 109.05 \\[0.1ex]
    Heteroscedastic MOCGP (HMOCGP) & \textbf{78.53} & \textbf{19.33} & \textbf{23.63} & 103.65\\[0.1ex]
    \hline
    \end{tabular}%
    \caption{Predictive performance across varying censoring overlap.}
\label{tab:overlap}%
\end{table}

\subsection{Copenhagen Dataset}
In this experiment, we are interested in modelling the true, latent demand of shared mobility services. 
In particular, our data comes from \emph{Donkey Republic}, a major bike-sharing provider in the city of Copenhagen, Denmark.
Given the presence of finite supply, censoring naturally arises every time the system is not able to satisfy all users requesting for service.
Because of this, being able to account for censored data is especially relevant in the planning and decision-making processes of shared transport modes, where the volatility of demand and the flexibility of supply modalities, require decisions to be made in strong accordance with user behavior and needs.
In this experiment, we use the multi-output GP formulation to exploit spatial correlations between nearby areas in the city of Copenhagen.
As in the case of the synthetic dataset, we would like to have access to the true demand process, free of any real world censorship. 
However, this ideal setting is impossible in the presence of real-world data, as historical demand records are intrinsically censored to some extent.
Thus, in our experiments, we assume the given historical data to represent the true demand (i.e. what ideally we would like to predict), to which we then artificially apply varying intensities of censoring.

\smallskip Concretely, we select $D=2$ nearby areas in the Donkey Republic network, for which we collect the daily time-series of bike-sharing demand, $\ystar_1(\bx)$ and $\ystar_2(\bx)$, which we assume to be Gaussian distributed.
For all those observations which we know to be censored (by analyzing the supply), we then compute $y_1(\bx) = (1-c) \, \ystar_1(\bx)$, where $c \in [0.2, 0.5, 0.8]$ is a given \emph{censoring intensity} which we use to examine the models' performance with varying severities of the censoring process, from slight to extreme, respectively.
In Table \ref{tab:donkey}, we can see prediction metrics evaluated through a k-fold cross-validation procedure.
As in the case of synthetic data, results show how the incremental flexibility allows the proposed model to achieve better performance in capturing the latent non-censored demand process. 
More importantly, the relative gap between the proposed HMOCGP and the other implemented models increases in scenarios characterized by higher values of the censoring intensity $c$, showing a consistent understanding of the non-censored demand process also under extreme censoring dynamics.
\smallskip
\begin{table}[t]
\centering
\setlength\tabcolsep{5pt}
\begin{tabular}{p{1.5cm} c c c c c c}
    \hline 
    \centering & \multicolumn{2}{c}{$c=0.2$} & \multicolumn{2}{c}{$c=0.5$} &
    \multicolumn{2}{c}{$c=0.8$}\\ [0.2ex]
     & $\text{\textbf{R}}^2$ & \textbf{MAE}
     & $\text{\textbf{R}}^2$ & \textbf{MAE}
     & $\text{\textbf{R}}^2$ & \textbf{MAE}\\ [0.2ex]
    \hline
    NCGP & 0.54 & 7.68 & 0.51 & 7.94 & 0.46 & 8.29 \\ [0.1ex]
    MONCGP & 0.54 & 7.69 & 0.52 & 7.85 & 0.52 & 7.76 \\[0.1ex]
    CGP & 0.56 & 7.50 & 0.55 & 7.55 & 0.55 & 7.53 \\ [0.1ex]
    HCGP & 0.57 & 7.41 & 0.56 & 7.50 & 0.56 & 7.51 \\[0.1ex]
    MOCGP & \textbf{0.61} & 6.99 & \textbf{0.60} & 7.08 & 0.58 & 7.33 \\[0.1ex]
    HMOCGP & \textbf{0.61} & \textbf{6.98} & \textbf{0.60} & \textbf{6.99} & \textbf{0.59} & \textbf{7.18}\\[0.1ex]
    \hline
    \end{tabular}
    \caption{Copenhagen Dataset Test Results.}
\label{tab:donkey}%
\end{table}

\subsection{New York City Dataset}

\begin{figure}[t]
\centering
\includegraphics[width=.98\columnwidth]{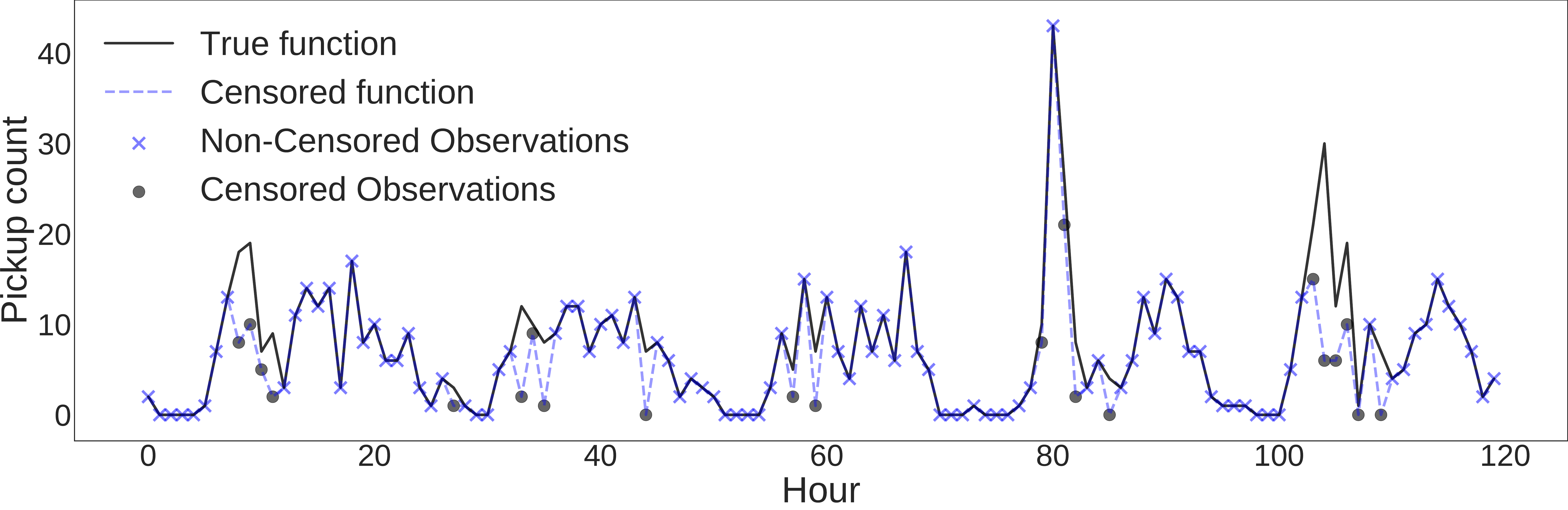}
\caption{Visual representation of the censoring process. 
Censoring limits the observability of the latent demand time-series (black, continuous line), by exclusively giving access to its censored counterpart (blue, dashed line).
In this example, censored models attempt to estimate the latent demand of mobility through a mix of censored (circles) and non-censored observations (crosses).}
\label{fig:censoring-example} 
\end{figure}

In our last experiment, we place special emphasis on the ability of the proposed model to generalize to other likelihood functions, namely Poisson and Negative Binomial, since the latter can be very useful when trying to infer the latent true demand of goods (i.e., typically count data). For that purpose, we consider the publicly-available NYC-CitiBike dataset \cite{CitiBike2020}. As in the previous experiment, the real non-censored demand is unavailable. Hence, for validating the proposed model, we proceed by assuming the observed demand to correspond to the true demand $\ystar_d(\bx)$ and manually censor it. Since this dataset does not provide information about the supply, we use the demand at previous time-steps as a proxy for determining when the supply may have reached zero - thus corresponding to a censored observation. This differs from the previous experiment, for which we had precise information about the supply and, therefore, the set of the censored observations $\mathcal{C}$ was known. Concretely, we assume the observation at time $t$ to be censored with probability $P(\bx_t \in \mathcal{C}) = 1-\mbox{Sigmoid}(\mbox{dropoffs}_{t-1} - \mbox{pickups}_{t-1} + 5)$. The intuition is that censoring is more likely to happen when the number of pickups is significantly higher than the number of dropoffs in the last time steps. Based on $P(\bx_t \in \mathcal{C})$, we then sample a set of censored observations $\mathcal{C}$, which we artificially censor by sampling: $y_d(\bx) \sim \mathcal{U}(0,\ystar_d(\bx))$. Figure~\ref{fig:censoring-example} shows an example of applying this procedure to one of the demand time-series considered. Given a train set with censored observations, the goal is to verify the ability of the different implemented models to correctly estimate the true demand for a test set. 

For this experiment, we consider four randomly selected locations: Station 3386 in Carroll Gardens, Station 2002 in Williamsburg, Station 3711 in East Village and Station 379 in Midtown Manhattan. We use 1 month of hourly observations for training, 10 days for validation (for early-stopping) and another 20 days for testing. For the multi-output variants, we jointly model each of the time-series with the most correlated time-series ($D=2$) in the data after censoring (i.e. both training time-series are censored). We parameterize the Poisson likelihood with the rate given by $\mbox{Softplus(\textbf{f})}$ in order to ensure positivity. As for the Negative Binomial likelihood, we consider the following parameterization of the PMF: 
\begin{align}
P(X=k) = \binom{k+r-1}{k} \, p^r \, (1-p)^k,
\end{align}
where $r$ is the number of failures until the experiment is stopped and $p$ is the probability of success. We then further parameterize the number of failures as $r=1/\alpha$ and the log-odds for the probabilities of success as $\mu \alpha$, and proceed by placing Gaussian process priors over the vectors $\boldsymbol\alpha$ and $\boldsymbol\mu$, using a softplus link function in order to ensure positivity whenever necessary. 

Table~\ref{tab:nyc} shows the obtained results for the four stations considered. For all of them, and regardless of the likelihood function considered, we can observe that the proposed HMOCGP (or MOCGP, for the Poisson case) approach outperforms its non-censored and non-multi-output counterparts. This again highlights the importance of exploiting correlations between multiple outputs to overcome the problem of censoring, and further supports our hypotheses that the proposed approach generalizes well to other assumptions regarding the distribution of the target variable. In fact, the obtained results further illustrate how violating the assumption that the residuals are Gaussian distributed can have a considerable impact on the model's ability to infer the true/non-censored values of the target variable and its corresponding predictive performance. This is the case for Station 3386, where assuming a Poisson likelihood leads to better overall predictive performance when compared to the Gaussian likelihood that is typically used in the censored regression literature. More importantly, these experiments highlight the generality of the proposed framework, which can be used with different likelihood functions depending on their suitability for the given application - e.g.~Exponential distribution for rate data, Beta distribution for continuous data constrained in $[0,1]$, etc.  



\smallskip
\begin{table}[h]
\centering
\begin{tabular}{p{3.125cm} c c c c}
    \hline 
    & St. 3386 & St. 2002 & St. 3711 & St. 379\\ [0.1ex]
    \hline
    Gaussian NCGP & 0.241 & 0.466 & 0.172 & 0.637 \\[0.1ex]
    Gaussian MONCGP & 0.244 & 0.473 & 0.190 & 0.653 \\[0.1ex]
    Gaussian CGP & 0.360 & 0.484 & 0.339 & 0.795\\[0.1ex]
    Gaussian HCGP & 0.362 & 0.477 & 0.366 & 0.795\\[0.1ex]
    Gaussian HMOCGP & \textbf{0.369} & \textbf{0.671} & \textbf{0.381} & \textbf{0.800} \\[0.1ex]
    \hline
    Poisson NCGP & 0.352 & 0.466 & 0.150 & 0.678\\[0.1ex]
    Poisson MONCGP & 0.348 & 0.446 & 0.218 & 0.679\\[0.1ex]
    Poisson CGP & 0.383 & 0.476 & 0.201 & 0.697\\[0.1ex]
    Poisson MOCGP & \textbf{0.391} & \textbf{0.478} & \textbf{0.262} & \textbf{0.698} \\[0.1ex]
    \hline
    NegBin NCGP & 0.182 & 0.425 & 0.150 & 0.5712\\[0.1ex]
    NegBin MONCGP & 0.284 & 0.461 & 0.170 & 0.691\\[0.1ex]
    NegBin CGP & 0.203 & 0.486 & 0.174 & 0.701\\[0.1ex]
    NegBin HMOCGP & \textbf{0.376} & \textbf{0.490} & \textbf{0.224} & \textbf{0.730} \\[0.1ex]
    \hline
    \end{tabular}
    \caption{New York Dataset Test $R^2$.}
\label{tab:nyc}%
\end{table}

\subsection{Heteroscedasticity and Censored Data}
In this subsection, we aim to draw further connections between the concepts of heteroscedastic regression and censored data modelling.
In particular, we attempt to give a qualitative understanding of the reasons why we suggest that allowing models to account for input-dependent noise conditions can be relevant in the context of censored data.
As reviewed in previous sections, the censored-Gaussian from Eq.~\ref{eq:likelihood} can effectively be considered as a mixture distribution of (i) a Gaussian PDF term, and (ii) a $1$-minus-Gaussian CDF term, where the mixture assignments are defined by the set of censored observations $\mathcal{C}$.

\begin{figure}[t]
\centering
\includegraphics[width=.75\columnwidth]{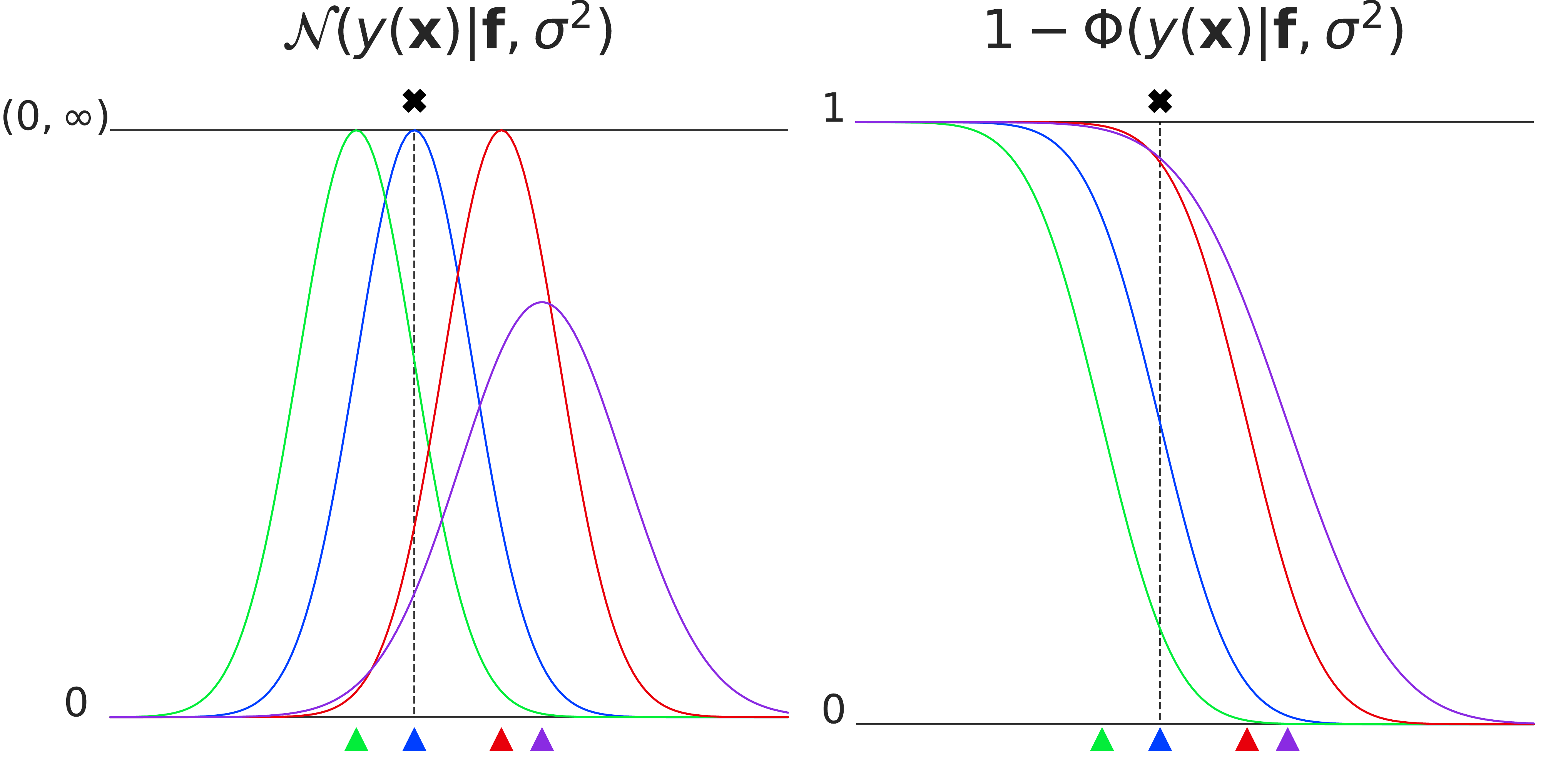}
\caption{Likelihood contributions for the censored-Gaussian distribution for either non-censored (left) or censored (right) observations evaluated in correspondence of the dashed vertical line.
Black crosses represent the assumed position for an observation $y(\bx)$, while the triangles correspond to four candidate mean values for $\boldf$. 
Through likelihood maximization, the plot shows the incentive for models to either \emph{exactly fit} the observed value in cases of non-censored data, or to \emph{over-estimate} it in case of censored data, where distributions over-estimating observed data correspond to higher likelihood values.}
\label{fig:hcg_intuition} 
\end{figure}

\smallskip In practice, Figure \ref{fig:hcg_intuition} shows a visualization of the likelihood contribution from the two mixture components for a given observation.
We note that, if the observation is censored (i.e. evaluated in the right plot), the purple Gaussian defined by higher variance is required to have a higher mean value $f(\bx)$ to obtain the same likelihood contribution of the red Gaussian.
Concretely, the variance parameter $\sigma$ directly controls the slope of the CDF function. 
For this reason, we suggest that, while constant noise would enforce the same amount of over-estimation for all observations, a heteroscedastic parameterization $\sigma(\bx)$ would allow the model to conditionally tune the amount of over-estimation required to better fit the unknown non-censored process. 

\section{Conclusions}
\label{sec:conclusions}
Building probabilistic models capable of dealing with censored data is instrumental in multiple research fields.
Recently, evidence has been gathered in favor of the combination of classical statistical approaches, such as Tobit models, with flexible model architectures (e.g., deep neural networks and Gaussian processes).
In this paper, we introduce a novel extension to the censored modeling toolbox based on three core observations. 
First, censored data sources often generate multiple correlated signals (e.g. demand of correlated goods limited by supply, multiple measurements of the same physical phenomenon, etc.). Thus, in order to better estimate the underlying non-censored signal, any statistical model should attempt to exploit these correlations. 
Second, heteroscedastic noise assumptions gain additional importance in the context of censored likelihood-based model estimation.
Third, current censored models fail to allow response variables to have generalizable error distributions specific for the problem at hand (e.g. Poisson for count data).

Based on these observations, in this paper, we introduce a novel heteroscedastic multi-output Gaussian processes capable of handling censored observations.
We show how HMOCPG is able to exploit information from correlated outputs, resulting in a better estimation of the underlying non-censored process.
We further extend the proposed framework to deal with both continuous and discrete outputs by using different likelihood functions.
We also draw connections between heteroscedastic regression and censored modelling, showing how the assumption of input-dependent noise can enable models to conditionally tune the over-estimation of censored observations in the case of censored likelihoods. 
Finally, given the resulting inference intractability, we derive a variational bound suitable for stochastic optimization.
Ablation studies on both synthetic and real world tasks show how the components characterizing the HMOCGP enable it to exploit information from correlated censored outputs, ultimately capturing complex censoring dynamics.

\smallskip In future work, we consider exploring different strategies for scaling the HMOCGP to very large datasets, such as sparse approximations \cite{HensmanEtAl2013} and amortized variational inference, in order to approximate the intractable posterior distribution over latent variables, \mbox{e.g.} as in \cite{LiuEtAl2019}. 
By avoiding the cubic complexity with respect to the number of observations, these approaches have the potential to extend the applicability of censored models to an even wider audience.

\bibliographystyle{plain}
\bibliography{paper3_bib}

\newpage
\appendix 
\section{Performance metrics}
\label{sec:app_performance}
We hereby report the definitions of the performance metrics used throughout this work.
Specifically, we will refer to (i) NLPD: Negative Log-Predictive Density, (ii) MAE: Mean Absolute Error, and (iii) $\text{R}^2$: coefficient of determination (or R-squared):
\begin{align}
        \text{NLPD} & =  - \sum_{i=1}^n \log p_{\theta}(y \, | \, \bx) \nonumber\\
        \text{MAE} & =  \frac{\sum_{i=1}^n |\hat{y_i} - y_i|}{n}\\
        \text{R}^2 & =  1 - \frac{\sum_{i=1}^n (y_i - \hat{y}_i)^2}{\sum_{i=1}^n (y_i - \overline{y})^2}, \hspace{5mm} \overline{y} = \frac{1}{n} \sum_{i=1}^n y_i, \nonumber
    \end{align}
    
where the following remarks are made in order: $p_{\theta}(\cdot)$ refers to the probability density function (or mass function, when considering discrete variables) defined by the learned GP model, $\hat{y}$ represents model predictions, $y$ and $\bx$ represent ground truth labels and input vectors respectively, and $n$ is the number of observations used in the evaluation.

\section{Synthetic Experiment}

\begin{figure*}[h!]
\begin{subfigure}{.5\textwidth}
\centering
\includegraphics[width=1\linewidth]{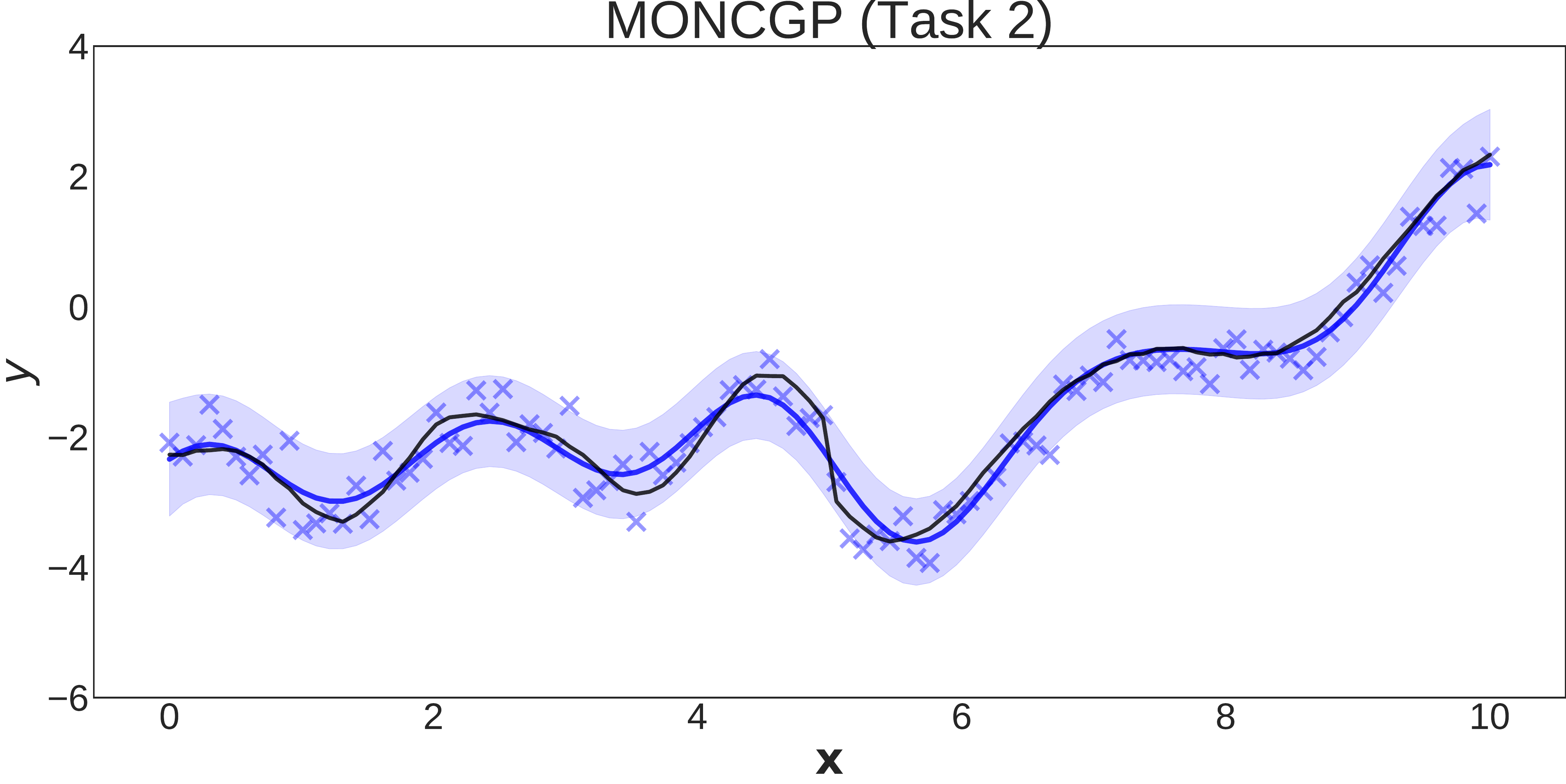}
\end{subfigure}
\hfill
\begin{subfigure}{.5\textwidth}
\centering
\includegraphics[width=1\linewidth]{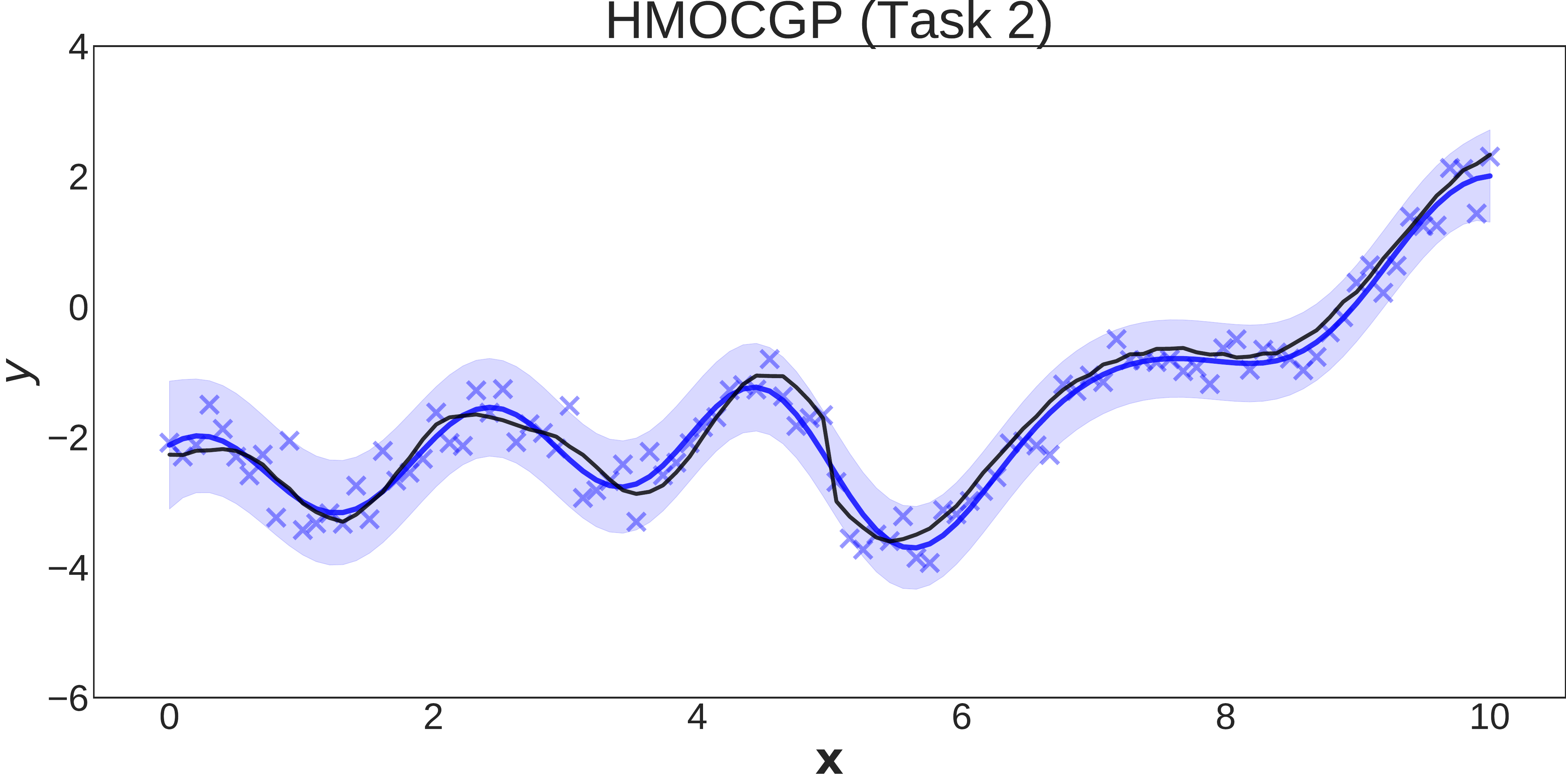}
\end{subfigure}
\caption{Comparison between MONCGP (left) and the proposed HMOCGP (right) for the second output $y_2(\bx)$ on synthetic data. Results show how both models are able to successfully fit the correlated output.}
\label{fig:synthetic_2}
\end{figure*}

\begin{figure*}[h!]
    \begin{subfigure}{.5\textwidth}
    \centering
    \includegraphics[width=1\linewidth]{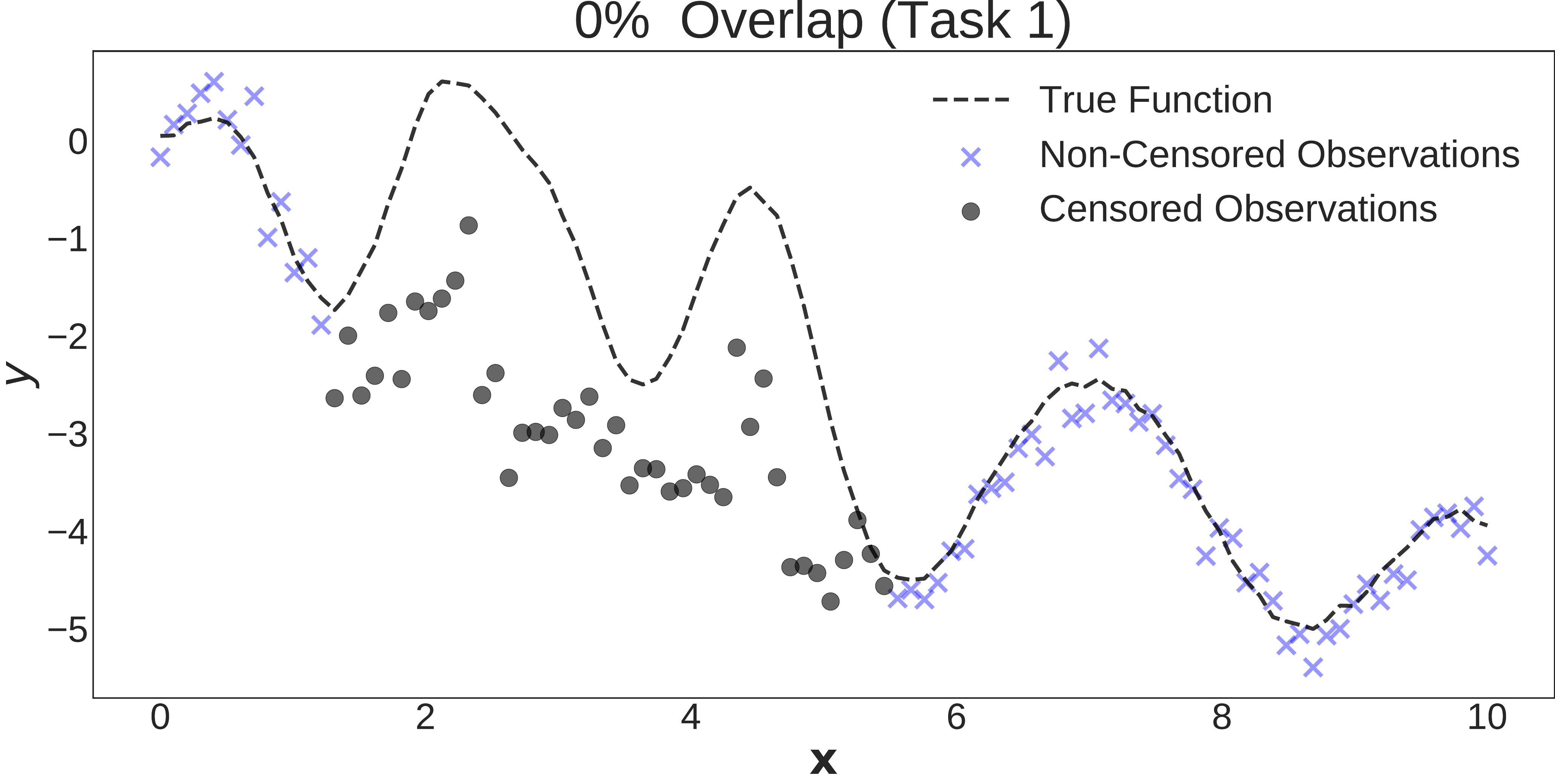}
    \end{subfigure}
    \hfill
    \begin{subfigure}{.5\textwidth}
    \centering
    \includegraphics[width=1\linewidth]{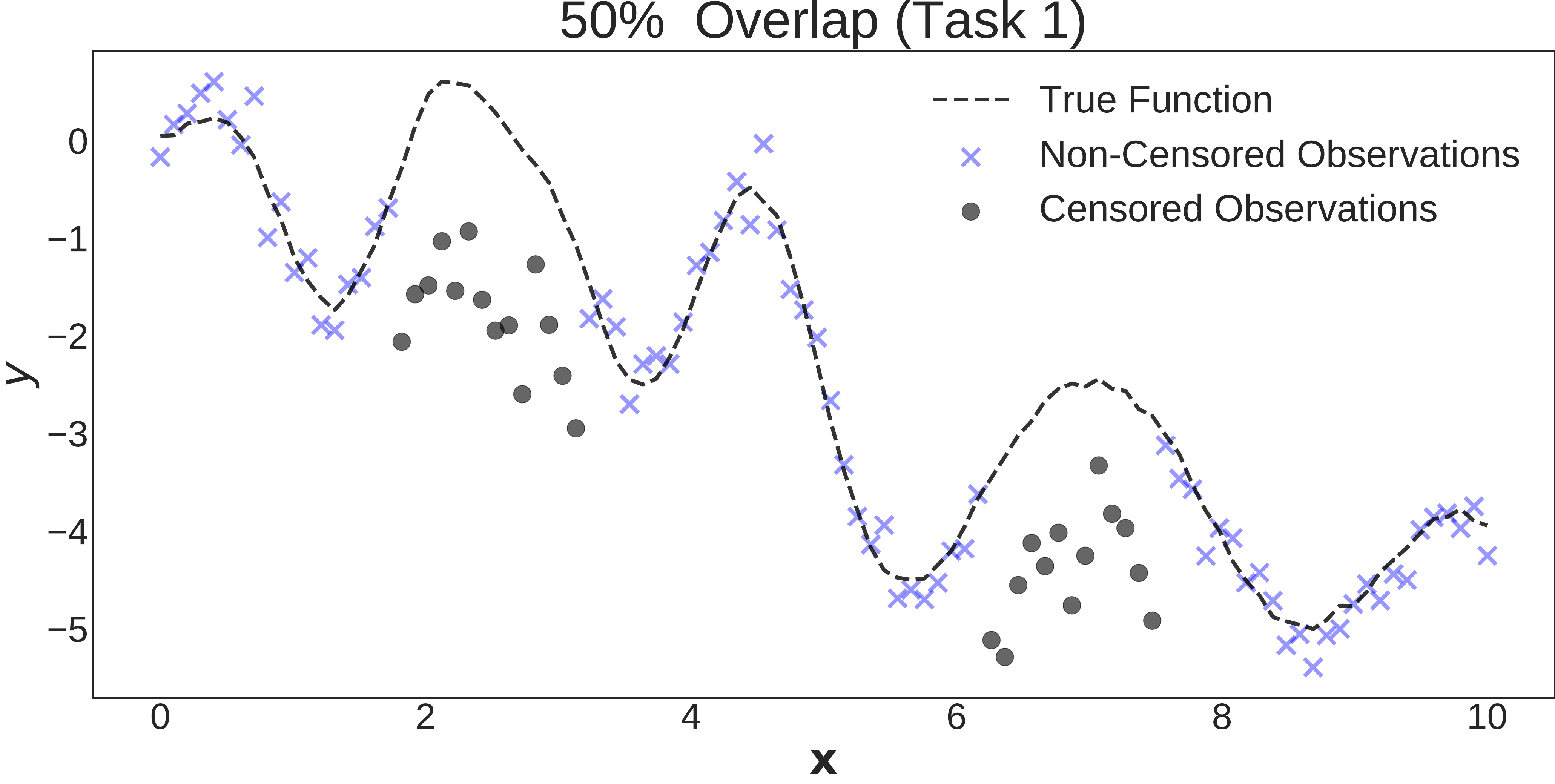}
    \end{subfigure}

    \begin{subfigure}{.5\textwidth}
    \centering
    \includegraphics[width=1\linewidth]{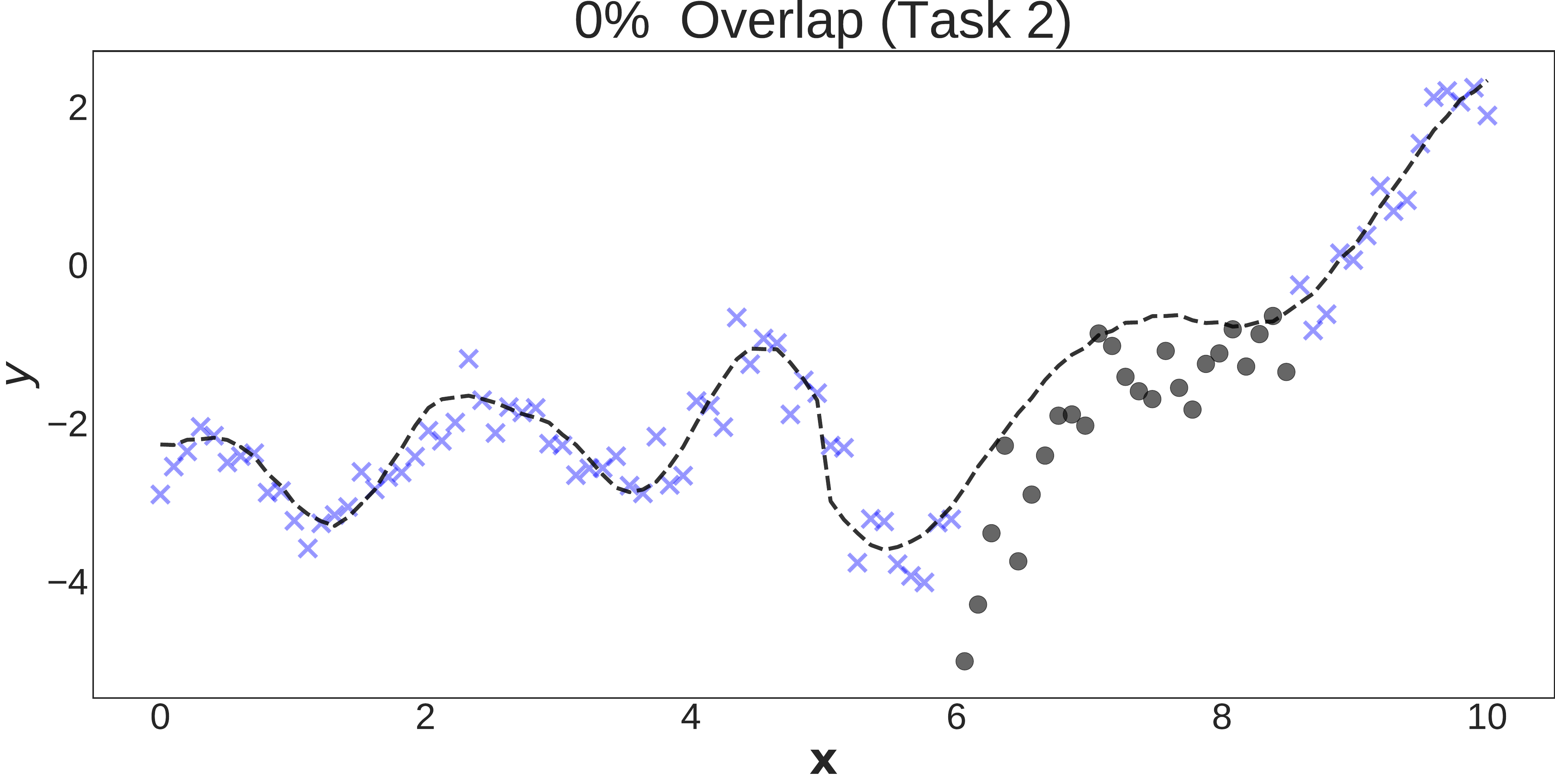}
    \end{subfigure}
    \begin{subfigure}{.5\textwidth}
    \centering
    \includegraphics[width=1\linewidth]{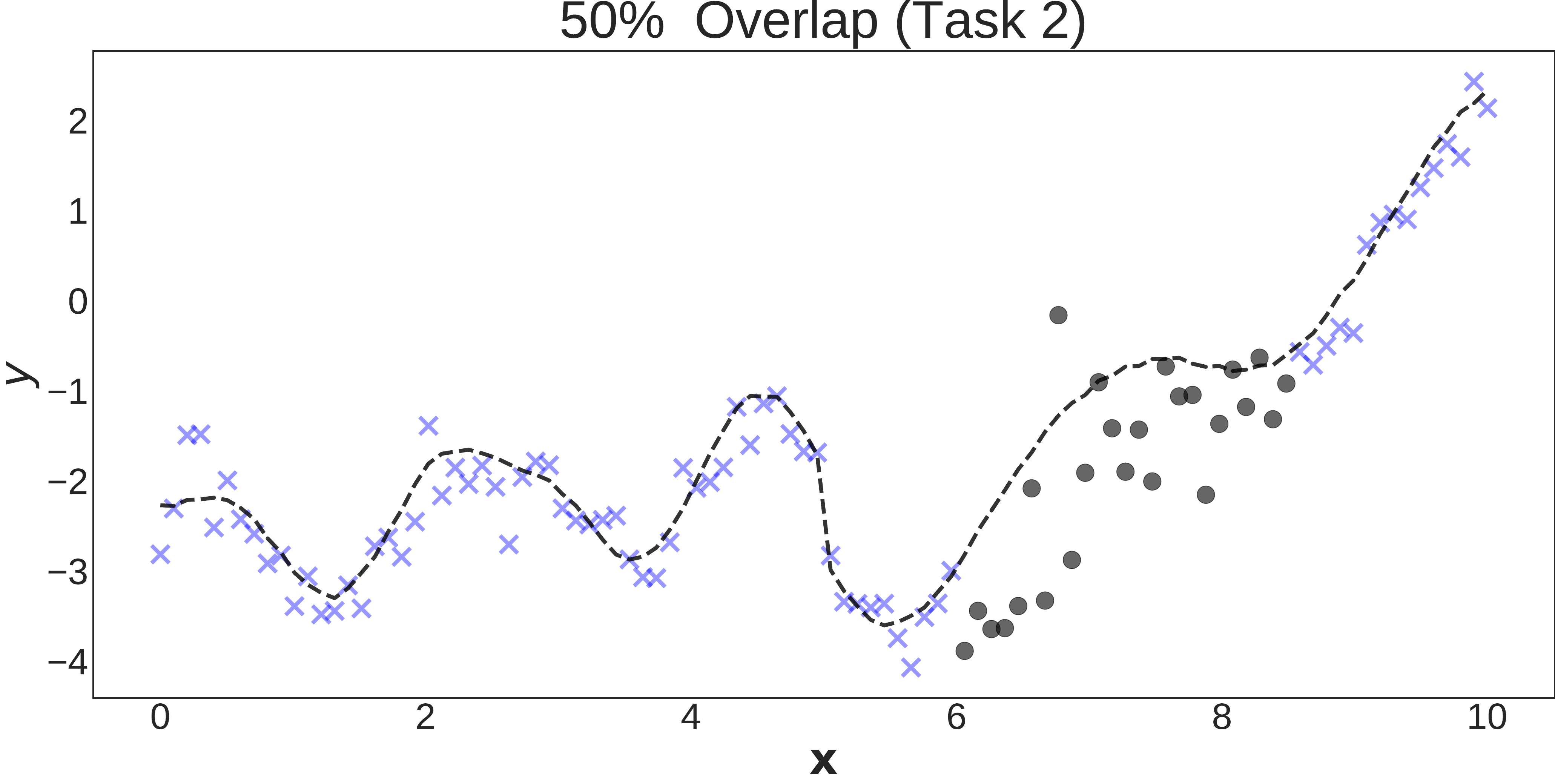}
    \end{subfigure}
    \caption{Visual representation of the synthetic experiment generated to analyse the effect of censoring overlap on performance. Plots show tasks $y_1(\bx)$ (top) and $y_2(\bx)$ (bottom) in case of $0\%$ (left) and $50\%$ (right) overlap, respectively.}
    \label{fig:overlap1}
\end{figure*}

\begin{figure*}[h!]
    \begin{subfigure}{.5\textwidth}
    \centering
    \includegraphics[width=1\linewidth]{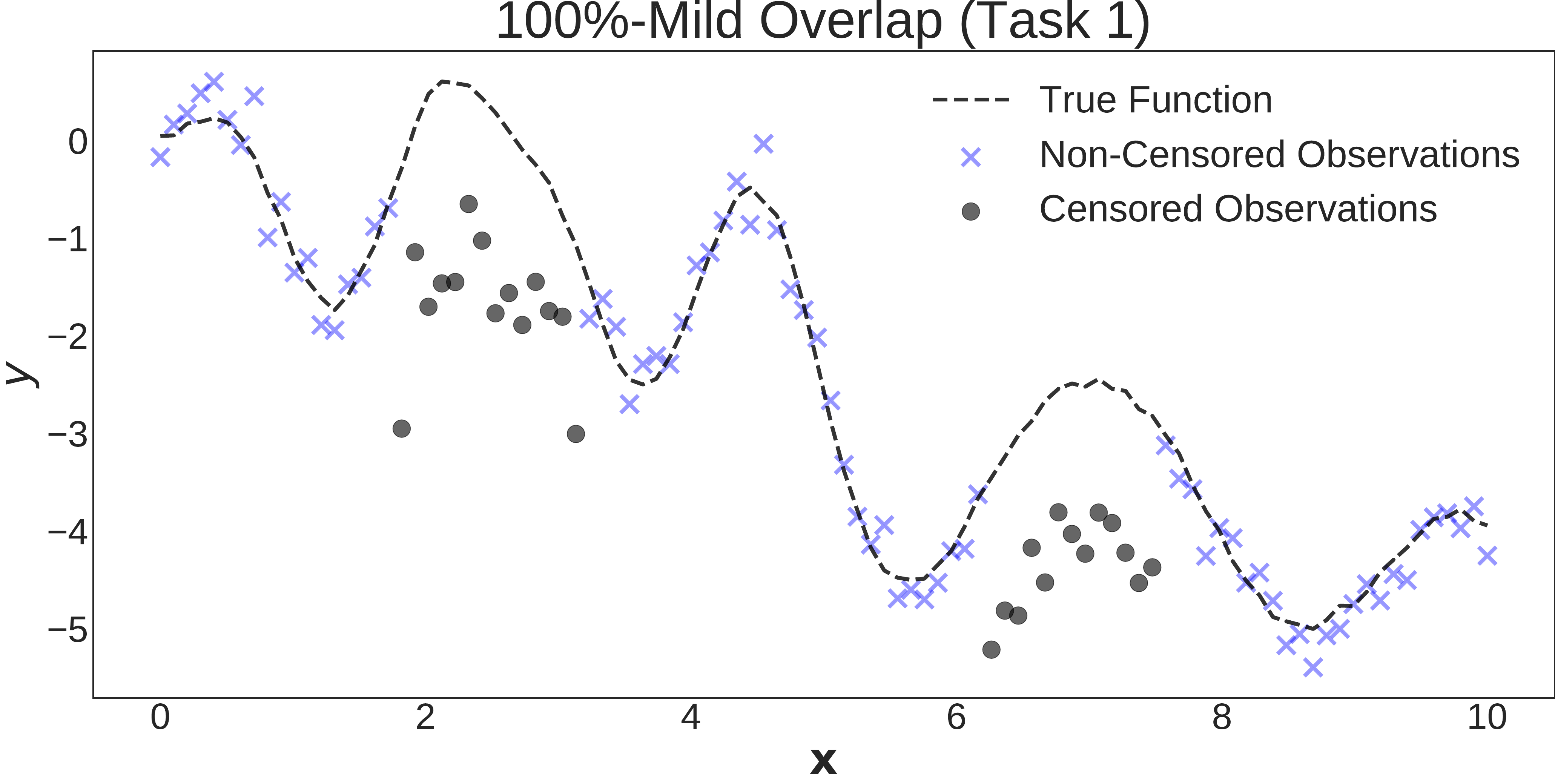}
    \end{subfigure}
    \hfill
    \begin{subfigure}{.5\textwidth}
    \centering
    \includegraphics[width=1\linewidth]{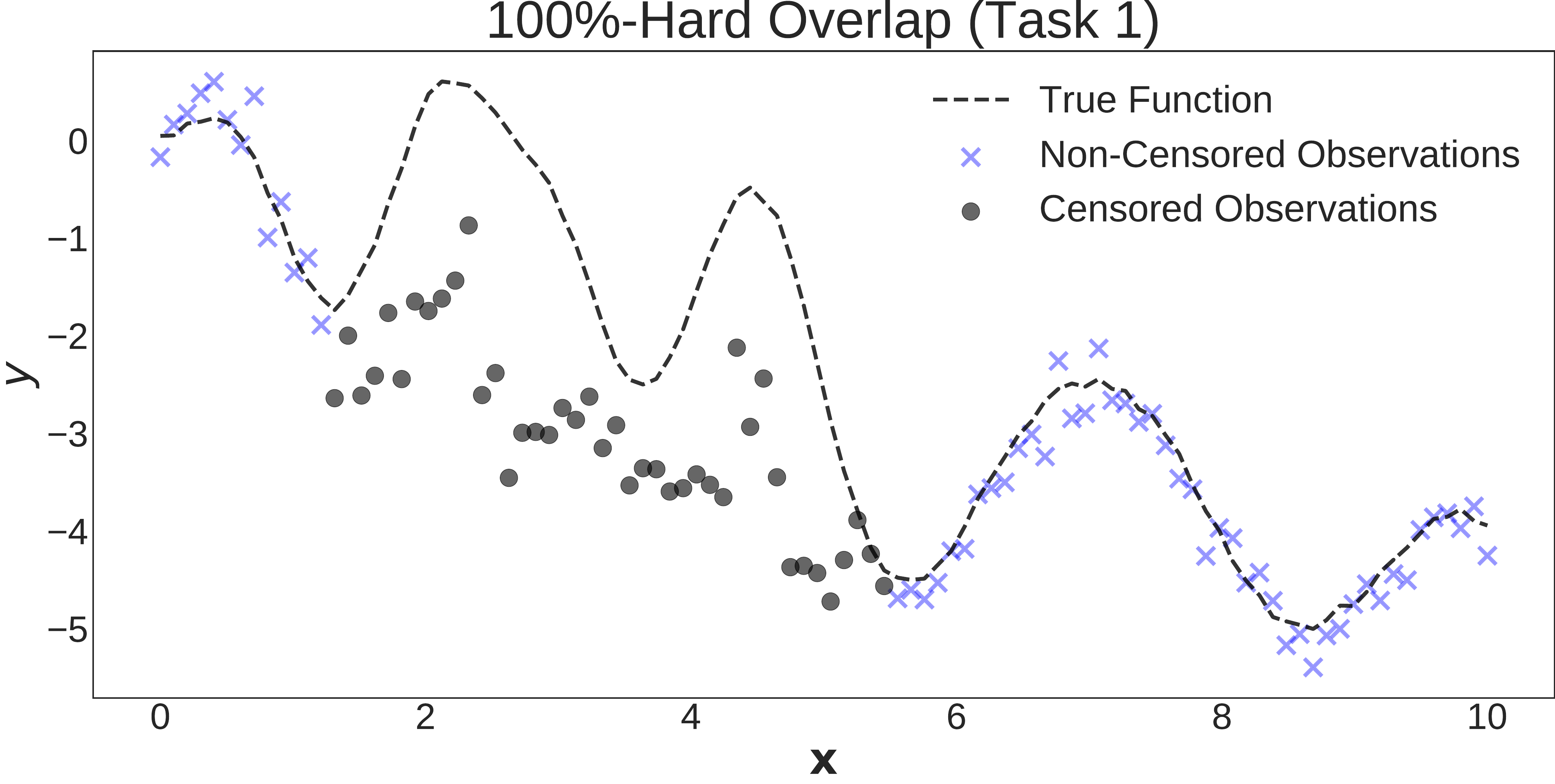}
    \end{subfigure}

    \begin{subfigure}{.5\textwidth}
    \centering
    \includegraphics[width=1\linewidth]{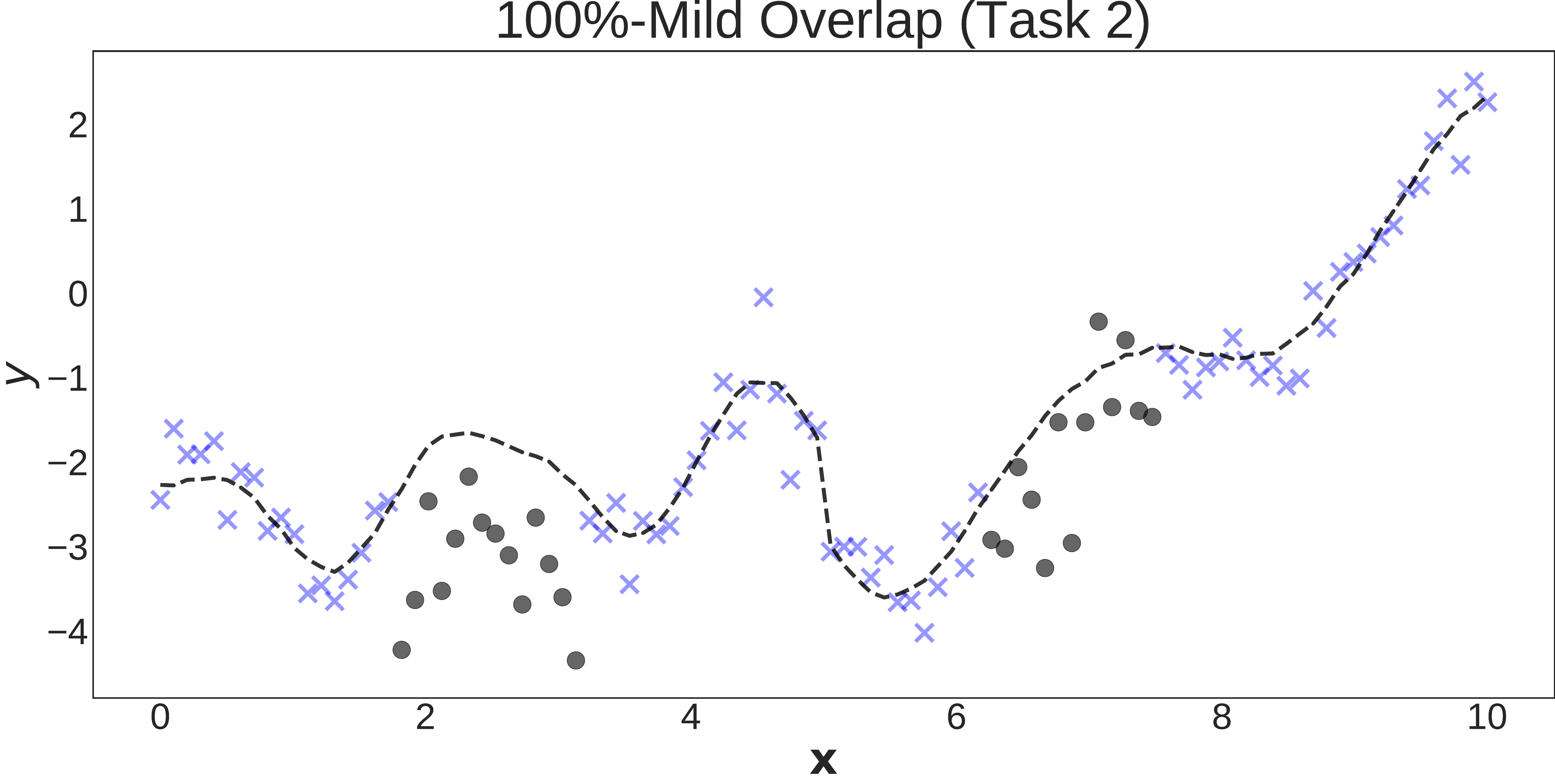}
    \end{subfigure}
    \begin{subfigure}{.5\textwidth}
    \centering
    \includegraphics[width=1\linewidth]{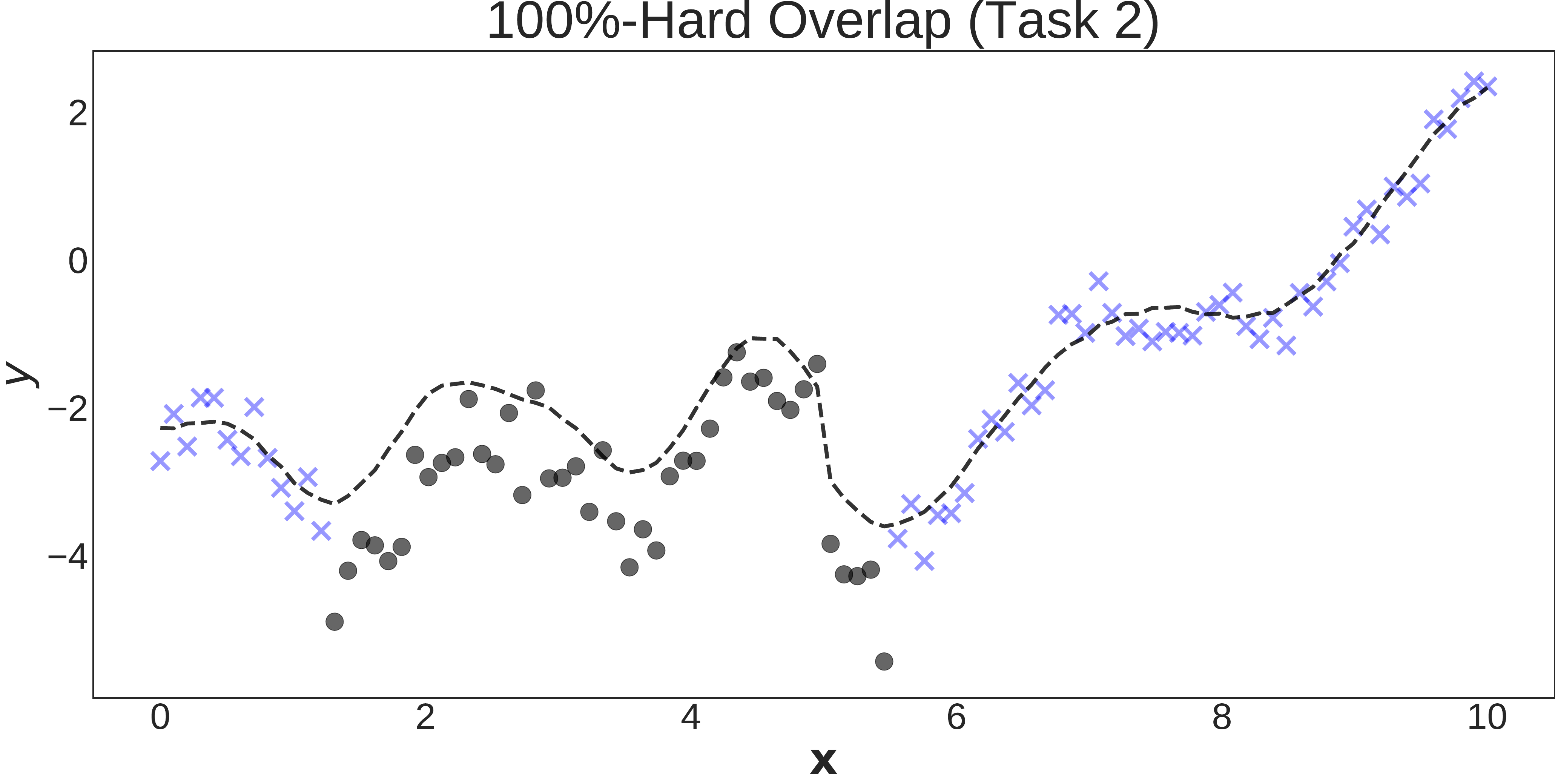}
    \end{subfigure}
    \caption{Visual representation of the synthetic experiment generated to analyse the effect of censoring overlap on performance. Plots show tasks $y_1(\bx)$ (top) and $y_2(\bx)$ (bottom) in case of $100\%$, discontinuous (left) and $100\%$ continuous (right) overlap, respectively.}
    \label{fig:overlap2}
    \end{figure*} 

\end{document}